\documentclass[]{fairmeta}

\usepackage{amsmath, amssymb, bm, bbm, mathrsfs}
\usepackage{siunitx}

\usepackage[utf8]{inputenc}
\usepackage[T1]{fontenc}
\usepackage{microtype}

\usepackage[dvipsnames,table]{xcolor}
\usepackage{soul}
\setul{0.5ex}{0.3ex}

\usepackage{booktabs, tabularx, threeparttable, multirow, array, colortbl, makecell}
\usepackage{tabularray}
\UseTblrLibrary{booktabs}

\usepackage{enumitem}
\setlist[enumerate]{noitemsep, topsep=0pt}
\setlist[itemize]{noitemsep, topsep=0pt}

\usepackage{caption}
\usepackage{subcaption}

\usepackage[ruled,vlined]{algorithm2e}

\usepackage{url}
\usepackage{pifont}
\usepackage{setspace}
\usepackage{wrapfig}
\usepackage{tikz}
\usetikzlibrary{positioning}
\usepackage[most]{tcolorbox}
\usepackage{wasysym}

\makeatletter
\newcommand{\thickhline}{%
    \noalign {\ifnum 0=`}\fi \hrule height 1pt
    \futurelet \reserved@a \@xhline
}
\makeatother

\usepackage{graphicx}
\usepackage{csquotes}
\usepackage{afterpage}
\usepackage{mathtools}
\usepackage{amsthm}
\usepackage{xspace}
\usepackage{nicefrac}
\usepackage{orcidlink}

\makeatletter
\DeclareRobustCommand\onedot{\futurelet\@let@token\@onedot}
\def\@onedot{\ifx\@let@token.\else.\null\fi\xspace}

\makeatother

\usepackage{orcidlink}

\definecolor{w_1}{RGB}{66,138,244}
\definecolor{w_2}{RGB}{73,144,245}
\definecolor{w_3}{RGB}{79,148,246}
\definecolor{w_4}{RGB}{87,154,247}
\definecolor{w_5}{RGB}{95,160,248}

\definecolor{uw}{RGB}{197,180,228}

\definecolor{fbApp}{HTML}{c8e7fa}
\definecolor{fbPurple3}{HTML}{f0ebf5}

\definecolor{citecolor}{HTML}{0071BC}
\definecolor{linkcolor}{HTML}{ED1C24}

\definecolor{citecolor}{HTML}{0071BC}
\definecolor{linkcolor}{HTML}{ED1C24}

\title{Language-Conditioned World Modeling for \\ Visual Navigation}

\author[1,*]{Yifei~Dong}
\author[1,*]{Fengyi~Wu}
\author[1]{Yilong~Dai}
\author[2]{Lingdong~Kong}
\author[1]{Guangyu~Chen}
\author[1]{Yetong~Sha}
\author[1]{Qiyu~Hu}
\author[3]{Feng~Liu}
\author[4]{Siyu~Huang}
\author[5]{Qi~Dai}
\author[1,\dagger]{Zhi-Qi~Cheng}

\affiliation[1]{University~of~Washington}
\affiliation[2]{National~University~of~Singapore}
\affiliation[3]{Drexel~University}
\affiliation[4]{Clemson~University}
\affiliation[5]{Microsoft~Research}

\contribution[*]{Equal contribution, listed in random order}
\contribution[\dagger]{Corresponding author.}

\abstract{
Goal-conditioned visual navigation has been a long-standing testbed for embodied AI. We study a natural language-conditioned variant, \textbf{language-conditioned visual navigation (LCVN)}, in which an embodied agent must follow a natural language instruction given only an initial egocentric observation. Without access to goal images, the agent must rely on language to shape its perception and continuous control. We introduce the LCVN Dataset, a benchmark of 39{,}016 trajectories and 117{,}048 human-verified instructions spanning diverse environments and instruction styles. Building on this benchmark, we study two complementary paradigms: (i) latent-imagination policy learning, in which a diffusion-based world model (LCVN-WM) imagines future observations and an actor--critic agent (LCVN-AC) learns its policy entirely within the imagined latent space; and (ii) unified autoregressive prediction, in which a single multimodal backbone (LCVN-Uni) jointly predicts actions and observations in one forward pass over a shared token sequence. Experiments show that two paradigms offer complementary strengths: latent imagination produces more temporally coherent rollouts, whereas unified prediction generalizes better to unseen environments. Targeted ablations further isolate the contributions of language guidance, conditioning signals, and instruction style, clarifying when language grounding versus dynamics modeling is the performance bottleneck. Together, these findings position LCVN as a testbed for studying how language, imagination, and decision-making interact in embodied agents.
  
}
  
\date{September 27, 2026}

\metadata[Code]{\url{https://github.com/UWMILab/LCVN}}

\metadata[Dataset]{\url{https://huggingface.co/datasets/fly1113/LCVN}}

\begin{document}

\maketitle

\section{Introduction}
\label{sec:intro}

\begin{figure*}[!t]

    \centering
    \includegraphics[width=0.85\linewidth]{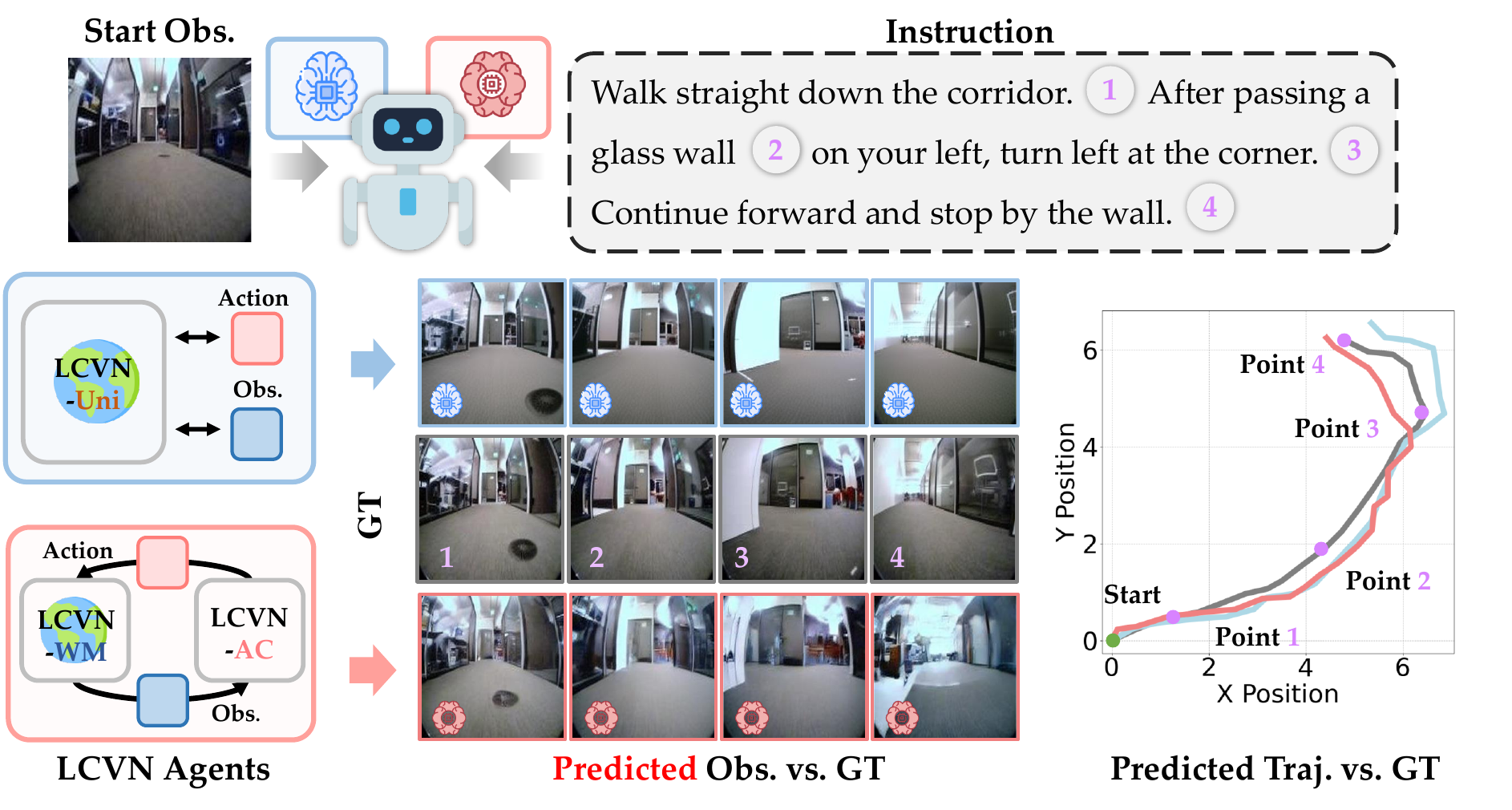}
    \caption{\small \textbf{Language-Conditioned Visual Navigation (LCVN).}~Given only an initial egocentric observation and a language instruction, the agent generates the entire future trajectory without environmental feedback, imagining intermediate states (\textcolor{violet}{\ding{192}}–\textcolor{violet}{\ding{195}}) along the described route. \textbf{LCVN-WM + LCVN-AC} imagines future latent observations and selects actions within the imagined latent space, while \textbf{LCVN-Uni} jointly predicts the next action and observation in a single autoregressive pass, conditioned on both visual and language inputs.
}
    \label{fig:task}
    
\end{figure*}

Goal-conditioned visual navigation has emerged as a central testbed for embodied AI, where an agent predicts a sequence of actions toward a specified goal from egocentric visual observations~\citep{sridhar2024nomad, shahvint, mirowski2016learning, chaplot2020learning, fu2022coupling, frey2023fast, pathak2018zero}. A dominant line of work specifies the goal as a visual image, a snapshot of the destination, and trains policies to predict trajectories that reach it~\citep{shahvint, sridhar2024nomad}. These goal-conditioned policies are typically trained on offline datasets of recorded egocentric trajectories and evaluated by their ability to predict future actions from a single starting observation, a protocol that has enabled large-scale data sharing across robots and environments~\citep{shah2022gnm}. 

Visual goals, however, can be a limiting interface for many real-world settings, since users rarely have a photograph of the intended destination at hand. A more natural and information-rich alternative is natural language, which can convey not only a target location but also intermediate landmarks, directional cues, and broader contextual constraints~\citep{zhang2022language, wu2025govig}. A brief instruction such as \emph{``After passing a glass wall on your left, turn left at the corner.''} often suffices to guide action even in unfamiliar places, and humans seamlessly integrate such guidance with what they see. Building embodied agents that can do the same, grounding language effectively in perception and continuous control, remains a core open challenge~\citep{google2024genie2, google2025genie3, survey_3d_4d_world_models, robosense_challenge_2025}.

A central difficulty is tightly coupling perception, prediction, and decision-making~\citep{worldlens,liang2026lidarcrafter,xu2025U4D,bian2025dynamiccity}. A promising route is to learn a world model as a neural simulator of environment dynamics, acquiring directly from data a predictive model of how observations evolve under actions~\citep{ha2018world, yao2025navmorph, worldlens, zhu2024sora}, which enables imagining future observations and planning over extended horizons. However, many existing designs~\citep{koh2021pathdreamer, bar2025navigation} focus on visual prediction while remaining largely decoupled from policy learning, limiting how agents adjust actions in response to linguistic guidance. Recent unified approaches~\citep{dong2025unified} mitigate this but suffer from inefficient interleaved training and inference. These limitations call for methods that better align language grounding, neural simulation, and policy learning.

To study this problem, we introduce \textbf{language-conditioned visual navigation (LCVN)} task (Fig.~\ref{fig:task}), in which an agent receives a natural language instruction and a single egocentric starting observation, and predicts the entire future trajectory toward the destination it describes. Following the trajectory-prediction protocol of prior goal-conditioned navigation work~\citep{shahvint, sridhar2024nomad, shah2022gnm}, training and evaluation operate on offline datasets of recorded trajectories, providing a reproducible testbed in which different solution approaches can be compared on identical inputs without dependence on interactive simulators. To support rigorous and systematic evaluation, we further construct \textbf{LCVN Dataset}, a large-scale corpus of language-annotated trajectories spanning diverse environments and instruction styles.


Building on this benchmark, we instantiate and systematically compare two paradigms that couple language grounding, future-state prediction, and policy learning in qualitatively different ways. The first paradigm (Fig.~\ref{fig:example}), latent-imagination policy learning, trains a language-conditioned diffusion world model \textbf{LCVN-WM} that imagines future latent observations, paired with an actor--critic agent \textbf{LCVN-AC} that learns its policy entirely within the imagined latent space via intrinsic rewards. The second paradigm (Fig.~\ref{fig:uni}), unified autoregressive prediction, fine-tunes a language-conditioned autoregressive world model \textbf{LCVN-Uni} that jointly predicts actions and observations in a single forward pass over a shared token sequence.

Extensive experiments show that LCVN agents outperform strong baselines on both navigation and imagination benchmarks, with latent-imagination policy learning producing more temporally coherent rollouts and unified autoregressive prediction generalizing better to unseen environments. Targeted ablations on language guidance, conditioning signals, and instruction style surface three findings that reframe what language-conditioned navigation actually requires: (i) language guidance functions as the goal specification that aligns states with destinations, supporting both navigation success and long-horizon rollout stability; (ii) primary bottleneck lies in grounding linguistic intent into continuous control rather than in language understanding itself; and (iii) navigation-relevant signal density matters more than descriptive richness. In summary, our contributions are:

\begin{enumerate}

    \item We formalize \textbf{language-conditioned visual navigation} task as a natural language variant of goal-conditioned navigation, and release the \textbf{LCVN Dataset}, a large-scale collection of 39{,}016 trajectories and 117{,}048 human-verified instructions in three styles (concise, intricate, landmark-based), assembled from five heterogeneous robotic platforms.

    \item We instantiate and systematically compare two paradigms on LCVN task: latent-imagination policy learning, in which a diffusion-based world model \textbf{LCVN-WM} imagines future visual states and an actor--critic agent \textbf{LCVN-AC} learns its policy entirely within the imagined latent space; and unified autoregressive prediction, in which a single multimodal backbone \textbf{LCVN-Uni} jointly predicts actions and observations in one forward pass.

    \item We provide experiments and ablations that characterize the behavior and generalization of both paradigms on the LCVN task, positioning LCVN as a shared testbed for studying how language, imagination, and decision making shape embodied behavior.

\end{enumerate}
\section{Related Work}
\label{sec:related work}

\noindent \textbf{Visual Navigation} spans goal-conditioned and free-exploration settings, with goal-conditioned navigation long regarded as a core robotics challenge~\citep{sridhar2024nomad, shahvint, mirowski2016learning, chaplot2020learning, fu2022coupling, frey2023fast, pathak2018zero}. In this setting, an agent receives context and target images and must generate trajectories to reach the goal~\citep{shahvint,sridhar2024nomad,deng2026anyimagenav}. Early systems combined neural SLAM with analytical planning~\citep{chaplot2020learning}, while later work learned direct mappings from egocentric observations to actions with reinforcement learning~\citep{chenlearning, lillicrap2015continuous, haarnoja2018soft}. More recent foundation-model approaches~\citep{shahvint, sridhar2024nomad, shah2022gnm} train policies on offline datasets of recorded trajectories to predict future actions from a single starting observation and a goal specification, a protocol that has enabled large-scale data sharing across heterogeneous robots and environments. However, these approaches typically specify the goal as a visual image, underutilizing the intent and contextual cues that language can convey~\citep{zhang2022language, wu2025govig}. In parallel, language-guided navigation has been studied in ObjectNav~\citep{habitatchallenge2023,deng2025mapbert}, VLN~\citep{anderson2018vision, krantz2020beyond}, and REVERIE~\citep{qi2020reverie}, which evaluate agents in simulators where observations are received at every step. LCVN occupies a distinct point: unlike discrete graphs~\citep{anderson2018vision, qi2020reverie} or simulator-bound continuous control~\citep{krantz2020beyond, habitatchallenge2023}, it inherits the offline trajectory-prediction protocol of goal-conditioned navigation with language goals over recorded real-robot trajectories, complementing closed-loop evaluation and scaling to heterogeneous real-world data across diverse platforms and environments.

\noindent \textbf{World Models in Navigation.} World models can be understood as learned neural simulators of environment dynamics~\citep{ding2024understanding, zhu2024sora,chu2026agentic}: rather than relying on hand-built physics or rendering engines, they acquire a predictive model of how observations evolve under actions directly from data. This perspective has driven a rapid evolution from recurrent latent dynamics~\citep{ha2018world, hafner2019dream, hafner2020mastering, hafner2024masteringdiversedomainsworld} to Transformer-based~\citep{assran2023self, bardes2024revisiting, karypidis2024dino, baldassarre2025back} and diffusion-based architectures~\citep{brooks2024video, agarwal2025cosmos, valevski2024diffusion, bar2025navigation, yu2025gamefactory}, with LLM-driven variants further modeling dynamics via language prompting~\citep{zhao2025drivedreamer, xing2025critiquesworldmodels, wang2025we}. Within navigation, two design philosophies have emerged. Policy-centric methods~\citep{shah2022gnm, shahvint, sridhar2024nomad} map observations directly to actions, treating future-state prediction as implicit. Model-centric methods explicitly predict future observations for planning, instantiated as GAN-based imagination in PathDreamer~\citep{koh2021pathdreamer}, diffusion video rollouts in NWM~\citep{bar2025navigation}, and unified autoregressive prediction in UniWM~\citep{dong2025unified}. Yet these works largely decouple imagination from policy learning: NWM omits language entirely, PathDreamer's auxiliary inputs hinder generalization~\citep{lin2024navcot}, and UniWM's interleaved training and inference are inefficient. How language grounding, neural simulation, and policy learning should be jointly designed under a single task remains underexplored. LCVN provides such a setting, in which we instantiate two paradigms: latent-imagination policy learning, pairing a diffusion-based world model LCVN-WM with an actor--critic agent LCVN-AC trained in its latent space; and unified autoregressive prediction, where a single multimodal backbone LCVN-Uni jointly predicts actions and observations.
\section{\textbf{LCVN}: Language-Conditioned Visual Navigation}
\label{sec:task}

In this section, we formalize the \textbf{language-conditioned visual navigation} task, a natural-language variant of goal-conditioned visual navigation. An embodied agent begins navigation with an egocentric RGB observation $o_s$ and receives a natural language instruction $I = \langle w_{1}, w_{2}, \ldots, w_{n} \rangle$, consisting of a sequence of words. The agent's goal is to generate a sequence of navigation actions $A_T = \{ \hat{a}_1, \hat{a}_2, \dots, \hat{a}_T \}$ that successfully guides it to the destination described by the instruction. Each action $\hat{a}_t$ can take the form of a continuous control command $(\mathbf{u}_t, \phi_t)$. In detail, the planar displacement $\mathbf{u}_t \in \mathbb{R}^2$ represents forward, backward, and lateral motion, together with a yaw rotation $\phi_t \in \mathbb{R}$. The action sequence terminates with a null action indicating arrival at the goal location. Following the trajectory-prediction protocol of prior goal-conditioned navigation work~\citep{shahvint, sridhar2024nomad, shah2022gnm}, the agent predicts this entire action sequence from the initial observation $o_s$ and the instruction $I$, and is evaluated on offline datasets of recorded trajectories by comparing the predicted trajectory against the recorded ground truth. The instruction $I$ serves as a persistent contextual condition that modulates the agent's policy throughout the episode.

\subsection{LCVN Dataset}
\label{sec:LCVN-dataset}

To support research in language-conditioned visual navigation, we construct \textbf{LCVN}, a large-scale multi-style instruction dataset built by augmenting five heterogeneous robotic trajectory datasets with human-verified language instructions. We use \textbf{Go Stanford}~\citep{hirose2018gonet}, \textbf{ReCon}~\citep{shah2021rapid}, \textbf{SCAND}~\citep{karnan2022socially}, and \textbf{HuRoN}~\citep{hirose2023sacson} for training and in-domain evaluation, covering complementary aspects of real-world navigation (e.g., open-world settings in ReCon, socially compliant behavior in SCAND), and reserve \textbf{TartanDrive}~\citep{triest2022tartandrive} as an unseen split for zero-shot evaluation, since its off-road driving scenes differ markedly from the indoor and structured outdoor environments seen in training, providing a strong distribution shift for testing generalization. Trajectories are standardized by normalizing per-frame displacement by each dataset's average step size, filtering out backward movements and trajectories shorter than three steps~\citep{bar2025navigation, sridhar2024nomad}, and segmented into semantically coherent sub-scenes using Qwen-3.5-9B~\citep{qwen35blog}. We adopt a seeded few-shot annotation strategy: our annotation team first manually writes a seed set of high-quality instructions, which then serves as the few-shot prompt context for Qwen-3.5-9B to generate instructions at scale~\citep{liu2025grl}, followed by human verification and refinement. Alongside the egocentric video, we additionally render a top-down 2D trajectory map as visual input, which disambiguates global heading changes from local motion noise. Each trajectory receives instructions in three styles: \textbf{Concise} (essential directional cues), \textbf{Intricate} (rich scene descriptions), and \textbf{Landmark-based} (anchored to environmental landmarks), enabling systematic evaluation of language grounding under varying levels of detail rather than a single uniform style~\citep{kolagar2024aligning}.

The final LCVN dataset contains 39,016 trajectories and 117,048 instructions, partitioned into training (28,813), val seen (3,602), val unseen (1,500, solely from TartanDrive), and test (5,101, from both TartanDrive and other sources). Beyond its primary application in language-conditioned visual navigation, LCVN dataset may also support related tasks like style-aware instruction generation~\citep{kong2024controllable, wu2025govig} and multimodal understanding~\citep{tong2025metamorph, zhang2025unified}. To verify annotation quality, a human evaluation on 640 randomly sampled trajectories yields an overall quality score of 5.2/6.0, confirming instruction quality across all three styles. Please kindly refer to \textbf{Appendix} for more dataset details.

\section{Paradigms for Language-Conditioned Visual Navigation}
\label{sec:LCVN-model}

\begin{figure*}[t]

\centering
\includegraphics[width=0.999\linewidth]{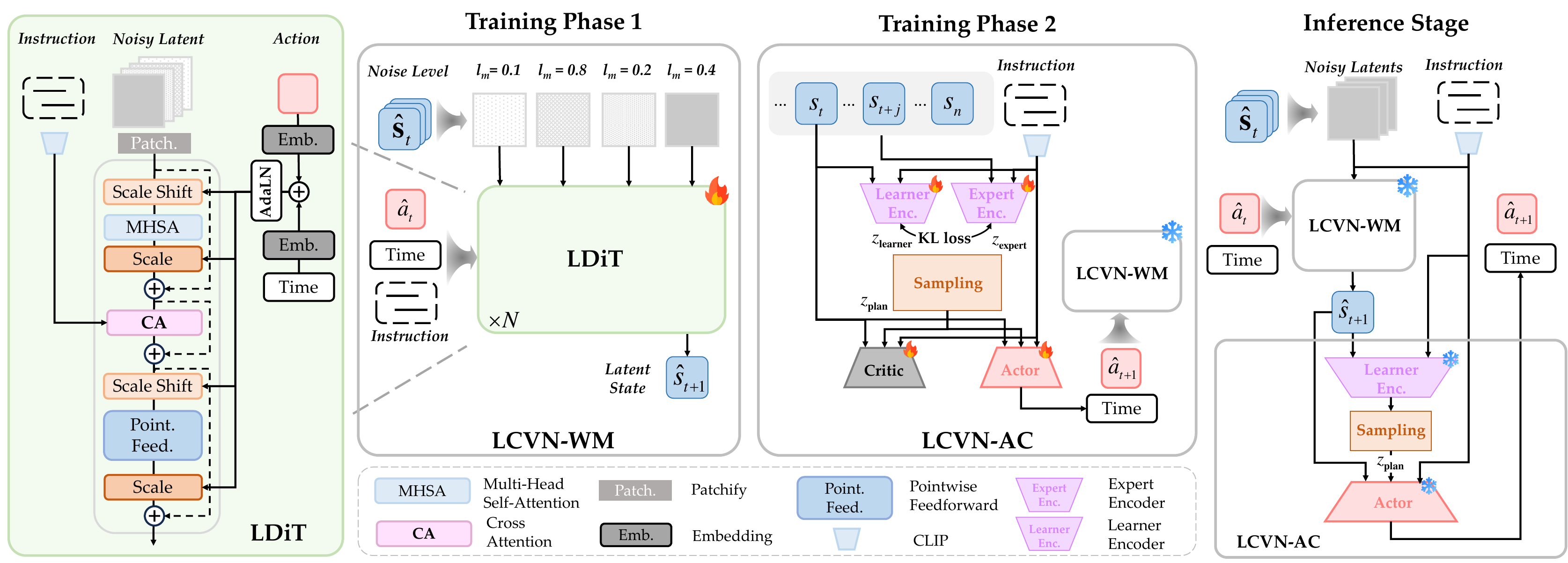}
\caption{\small
\textbf{Language-conditioned world model (LCVN-WM) and associated LCVN-AC.}
\textbf{Phase~1:} LCVN-WM is trained with Diffusion Forcing~\citep{chen2024diffusion, song2025history} to predict future latent observations from noisy context latents at independent noise levels, conditioned on actions, instruction, and time shift.
\textbf{Phase~2:} LCVN-AC is trained in LCVN-WM's latent space, aligning expert and learner plans via KL divergence and optimizing an actor--critic objective with intrinsic rewards that measure agreement between predicted and expert latent rollouts.
\textbf{Inference:} LCVN-WM predicts the next latent observation, on which LCVN-AC conditions to generate the next action.
}
\label{fig:example}

\end{figure*}

To solve the LCVN task, we instantiate and compare two paradigms that couple language grounding, future-state prediction, and policy learning in qualitatively different ways, under identical training data to enable a controlled comparison. Paradigm 1: Latent-imagination policy learning (Sec.~\ref{sec:LCVN-wm}--\ref{sec:LCVN-ac}, Fig.~\ref{fig:example}) trains a language-conditioned diffusion world model \textbf{LCVN-WM} that imagines future latent observations, paired with an actor--critic agent \textbf{LCVN-AC} that learns its policy within this latent space via intrinsic rollout rewards. Paradigm 2: Unified autoregressive prediction (Sec.~\ref{sec:LCVN-uni}, Fig.~\ref{fig:uni}) fine-tunes a language-conditioned autoregressive world model \textbf{LCVN-Uni} that jointly predicts the next action and observation in a single forward pass over a shared token sequence.

\subsection{LCVN-WM}
\label{sec:LCVN-wm}
LCVN-WM is the imagination component of Paradigm 1: it predicts future latent observations conditioned on past observations, the current action, and the language instruction, providing the imagined trajectories on which LCVN-AC's policy is trained. Architecturally, LCVN-WM extends the Diffusion Transformer (DiT)~\citep{peebles2023scalable} with language conditioning (termed LDiT) and employs diffusion forcing (DF)~\citep{chen2024diffusion, song2025history} to enable flexible noising patterns, illustrated in Fig.~\ref{fig:example}.

Given current observations encoded as VAE latents, LCVN-WM $F_{\theta}$ conditions on navigation action $\hat{a}_t$ where $t$ denotes timestep, instruction $I$, and time shift $t_s$ to predict the next state $\hat{s}_{t+1}$:
\begin{equation} 
    s_i\!=\!\text{Enc}_{\theta}(o_{i})~~~
    \hat{s}_{t+1}
    \!\sim\! F_{\theta}\big(\hat{s}_{t+1} \mid \hat{\mathbf{s}}_t,\, \hat{a}_{t}, I, t_s\big), 
\label{eq:LCVN-wm} 
\end{equation} 
where $o_i$ denotes image observation and is encoded by a learned VAE encoder to form $s_i$, $\hat{\mathbf{s}}_t = [\hat{s}_t,\hat{s}_{t-1},..., \hat{s}_{t-k}]$ represents the previous estimate of the state, and $k$ is the context size.

\noindent\textbf{Language Conditioning.}~To enable language conditioning, we encode the instruction $I$ into $I_{clip}$ using a frozen CLIP text encoder~\citep{radford2021learning}. 
The LDiT architecture integrates a multi-head cross-attention layer within each DiT block, positioned between the self-attention and feed-forward layers, as shown on the right side of Fig.~\ref{fig:example}. This design enables flexible interaction between visual latents and the instruction embeddings $I_{clip}$, which are directly fed into the cross-attention mechanism, thereby promoting semantic consistency between states and goals.

Meanwhile, we also include action and time conditioning within the conditioning mechanism following the settings in~\citep{bar2025navigation}. The navigation action $\hat{a}_t$ is encoded by first applying sine–cosine positional encoding to each scalar, followed by a MLP, and then concatenating results into a single vector. A similar procedure is applied to encode relative timeshift $t_s$ and diffusion timestep. The final vector is obtained by summing all three embeddings, which is then used in an AdaLN layer to generate scale and shift parameters that modulate outputs of Layer Normalization and intermediate activations in each attention block.  

\noindent\textbf{Diffusion Forcing.}~LCVN-WM adopts the DF paradigm to strengthen temporal modeling in long-horizon navigation. Unlike standard diffusion training, DF applies independent noise levels $\ell_m$ to each latent state $\hat{s}_{t-m}$ within the context sequence $\hat{\mathbf{s}}_t = [\hat{s}_t, \hat{s}_{t-1}, \ldots, \hat{s}_{t-k}]$:  
\begin{equation}
\hat{s}_{t-m}^{(\ell_m)} = \sqrt{\alpha_{\ell_m}} \,\hat{s}_{t-m} + \sqrt{1 - \alpha_{\ell_m}} \,\epsilon_m,
\label{eq:df}
\end{equation}
where $\epsilon_m \sim \mathcal{N}(0, I)$ and $\alpha_{\ell_m}$ follow the standard noise schedule. DF introduces heterogeneous corruption across the context window, encouraging stronger temporal modeling~\citep{chen2024diffusion, song2025history}.

\subsection{LCVN-AC}
\label{sec:LCVN-ac} 

LCVN-AC is the policy component of Paradigm 1: while LCVN-WM imagines latent rollouts, LCVN-AC learns a language-conditioned policy that selects actions to follow the instruction, trained entirely within LCVN-WM's latent space (Fig.~\ref{fig:example}). Concretely, the agent learns a policy $\pi_{\theta}(\hat{a}_{t+1} \mid {s}_t, I_{clip})$ and a value function $v_\psi({s}_t, I_{clip})$ from the LCVN dataset, where $\pi_{\theta}$ denotes the actor, $v_\psi$ the critic, and $I_{clip}$ the CLIP-encoded instruction embedding.

\noindent \textbf{Trajectory Encoding and Language Conditioning.}~Following \citep{mees2022matters, nematollahi25icra}, we replace high-dimensional RGB images with compact world-model latents. In addition to encoding first-person observations, we represent a latent trajectory segment from the current state to the goal, consisting of continuous navigation actions $a \in \mathbb{R}^3$ across timesteps, and compress each trajectory into a latent plan $z_{\text{plan}}$ using a seq2seq CVAE~\citep{lynch2020learning, xie2025latent}. During training, an expert encoder $q_{\text{expert}}$ consumes the full trajectory to produce expert plan $z_{\text{expert}}$, while a learner encoder $q_{\text{learner}}$ maps $({s}_t, I_{clip})$ at test time to predict $z_{\text{plan}}$. We align these encoders by minimizing the KL divergence:
\begin{equation}
\mathcal{L}_{\text{KL}} = D_{\text{KL}}\Big( 
  q_{\text{expert}}(z_{\text{expert}} \mid s_{t:n}, I_{clip})
  \,\|\, q_{\text{learner}}(z_{\text{plan}} \mid s_t, I_{clip}) \Big),
\label{eq:kl}
\end{equation}
where $n$ denotes total trajectory length. Aligning expert and learner encoders with instruction embeddings improves efficiency and enforces semantic consistency between states and goals, ensuring latent plans remain grounded in the intended destination. For language conditioning, we maximize cosine similarity between LCVN-WM latents and corresponding instruction embeddings while minimizing it for mismatched pairs~\citep{radford2021learning, nematollahi25icra}, resulting in loss $\mathcal{L}_{\text{ins}}$.

\noindent \textbf{Actor–Critic with Intrinsic Reward.}~At each step, we sample a latent plan $z_{\text{plan}} \sim q_{\text{learner}}(\cdot \mid {s}_t, I_{clip})$, then use actor $\pi_\theta$ to generate action $\hat{a}_t \sim \pi_\theta(\cdot \mid {s}_t, I_{clip}, z_{\text{plan}})$. The next latent state is predicted by frozen LCVN-WM $F_{\theta}$:
\begin{equation}
\hat{s}^\pi_{t+1} \sim F_{\theta}(\hat{s}^\pi_{t+1} \mid {\mathbf{s}}_t, \hat{a}_t, I_{clip}, t_s),
\label{eq:ac_reward_wm}
\end{equation}
where ${\mathbf{s}}_t = [{s}_t,{s}_{t-1},..., {s}_{t-k}]$ denotes the sequence of previous latent states and $k$ is the context size. 
For the initial step ($t=0$), when past states ${s}_{t-1}, \ldots, {s}_{t-k}$ are unavailable, we pad the history by repeating the initial latent ${s}_s$. To mitigate covariate shift, we compute an intrinsic reward~\citep{demoss2023ditto} that measures the alignment between predicted and expert trajectories in latent space:
\begin{equation}
r^{\text{int}}_{t}({s}^E_{t+1}, \hat{s}^\pi_{t+1})
= \frac{{s}^E_{t+1} \cdot \hat{s}^\pi_{t+1}}{\max(\lVert {s}^E_{t+1}\rVert, \lVert \hat{s}^\pi_{t+1}\rVert)^2}~,
\label{eq:ac-reward}
\end{equation}
where ${s}^E_{t+1}$ is expert latent state at step $t+1$. The critic $v_\psi$ estimates the expected discounted return by:
\begin{equation}
    v_\psi({s}_t, I_{clip}, z_{\text{plan}}) \approx
    \mathbb{E}_{\pi_\theta, p_\phi} \!\left[ \sum_{k=t}^{t+H-1}\nolimits \gamma^{k - t} r^{\text{int}}_k \right],
\label{eq:ac-return}
\end{equation}
where $\gamma \in [0,1]$ is the discount factor and $H$ is the rollout horizon. This formulation allows the critic to estimate the cumulative intrinsic reward under the current policy $\pi_\theta$ and latent plan $z_{\text{plan}}$. We repeat this rollout over the entire trajectory horizon to produce $\{\hat{s}^\pi_k\}_{k=t+1}^{n}$, and compute intrinsic rewards and discounted returns for critic supervision.   

During training, the critic regresses to $\lambda$-return~\citep{hafner2020mastering}:
\begin{equation} 
\mathcal{L}(\psi)=\mathbb{E}_{\pi_\theta, p_\phi}\left[\sum_{k=t}^{t+H-1}\nolimits \frac{1}{2}\left(v_\psi(\hat{s}_k, I_{clip}, z_{plan})-\mathrm{sg}(V^\lambda_k)\right)^2\right],
\label{eq:ac-crtic}
\end{equation} 
where $\operatorname{sg}(\cdot)$ is the stop-gradient operator, $p_\phi$ denotes transition dynamics of LCVN-WM, and $V^\lambda_k$ represents $\lambda$
-return computed with a slowly updated target critic $v_\psi$. On the other hand, the actor maximizes expected return while regularizing plan consistency and language alignment:
\begin{equation}
\!\!\!\!\!\!\mathcal{L}(\theta)\!=\!
\mathbb{E}_{\pi_\theta, p_\phi}\!\!\left[
\!\sum\nolimits_{k=t}^{t\!+\!H\!-\!1} \!\!\big(\!\!-\!\! V^\lambda_k
\!\!+\!\!\alpha_1 \mathcal{L}_{\text{KL}}^k
\!\!+\!\!\alpha_2 \mathcal{L}_{\text{ins}}^k
\big)
\!\right]\!.
\label{eq:ac-actor}
\end{equation}
At inference time, agent plans next action based on current latent state $\hat{s}_{t+1}$ predicted by LCVN-WM, together with instruction embedding $I_{clip}$ and $z_{\text{plan}}$:
\begin{equation} 
\hat{a}_{t+1} \sim \pi_\theta(\hat{a}_{t+1} \mid {s}_{t+1}, I_{clip}, z_{\text{plan}})~.
\label{eq:LCVN-ac} 
\end{equation}

\subsection{LCVN-Uni}

\label{sec:LCVN-uni}
\begin{wrapfigure}{r}{0.45\textwidth}

\centering
\includegraphics[width=0.45\textwidth]{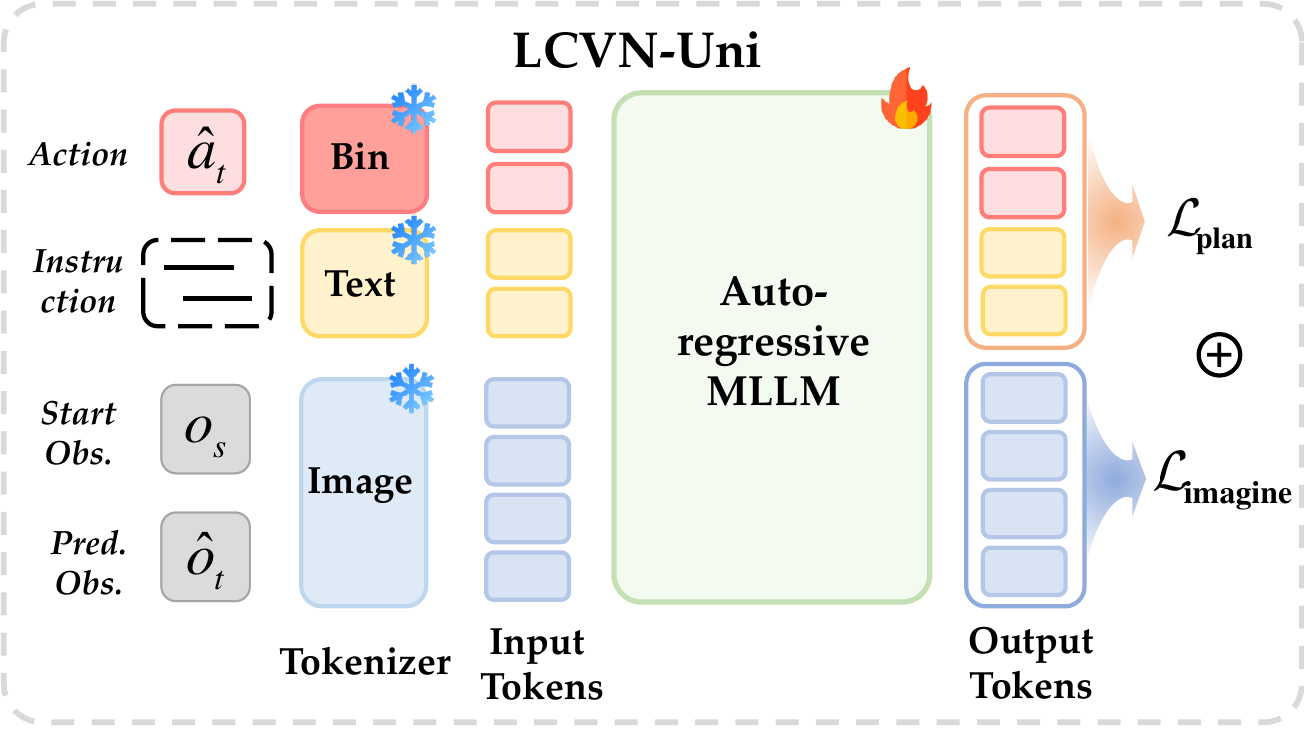}
\caption{\small \textbf{LCVN-Uni architecture.}~LCVN-Uni unifies navigation planning and world modeling within an autoregressive MLLM backbone.~Actions $\hat{a}_t$, instructions $I$, and observations $o_s, \hat{o}_t$ are tokenized by bin, BPE, and VQ tokenizers, respectively, then fused into a unified sequence.}
\label{fig:uni}
\vspace{-0.6cm}
\end{wrapfigure}

LCVN-Uni instantiates Paradigm 2: instead of separating an explicit world model from a downstream policy, it folds imagination and planning into a single multimodal autoregressive backbone $G_{\theta}$ (Fig.~\ref{fig:uni}) that produces the next action and the next observation in one forward pass:
\begin{equation} 
(\hat{a}_{t+1},\, \hat{o}_{t+1}) 
\;=\; 
G_{\theta}\big(\hat{o}_t,\, \hat{a}_t,\, I ,\,o_s \big). 
\label{eq:LCVN-uni} 
\end{equation} 
Architecturally, LCVN-Uni builds on~\citep{dong2025unified,team2024chameleon}, but trains action and observation prediction jointly rather than alternately, sharing representations between planning and imagination at every step.

\noindent \textbf{Data Preprocessing.}~Each training sample consists of $(o_t, a_t, o_s, I)$ paired with two targets, the next action $\hat{a}_{t+1}$ and the next observation $\hat{o}_{t+1}$ (see \textbf{Appendix} for prompt design). This eliminates need for separate sample types for planning and world modeling, and streamlines the training pipeline.

\noindent \textbf{Multimodal Tokenization.}~We employ three tokenizers~\citep{dong2025unified} to couple multimodal inputs: a VQ image tokenizer~\citep{gafni2022make} for visual observations $o_t$ and $o_s$, a BPE tokenizer~\citep{team2024chameleon} for instructions $I$ and text prompts, and a bin tokenizer~\citep{dong2025unified} for actions $a_t$. The resulting token sequences are fed to a causal Transformer that jointly models planning and imagination.

\noindent \textbf{Joint Training Objective.}~At each training step, the model produces logits for both action bins and visual tokens, enabling end-to-end joint optimization:
\begin{equation}
\mathcal{L}_{\text{joint}} = \mathcal{L}_{\text{plan}} + \lambda \mathcal{L}_{\text{imagine}},
\end{equation}
where $\mathcal{L}_{\text{plan}}$ is the discretized bin token loss~\citep{dong2025unified} for action prediction, $\mathcal{L}_{\text{imagine}}$~\citep{dong2025unified, li2025imagine} is the reconstruction loss for predicted future observations, and $\lambda$ balances the two. This shared objective lets the same backbone learn representations that serve both planning and visual prediction.

\noindent \textbf{Inference.}~At test time, LCVN-Uni generates the next action and observation in a single forward pass, avoiding alternating substeps used in~\citep{dong2025unified}. The predicted observation $\hat{o}_{t+1}$ is fed back as input observation at next step, so the agent unrolls its trajectory entirely in its own imagined visual stream.

\section{Experiments}

\label{sec: experiments}
\begin{figure*}[t]
 
    \centering    
   \includegraphics[width=0.999\linewidth]{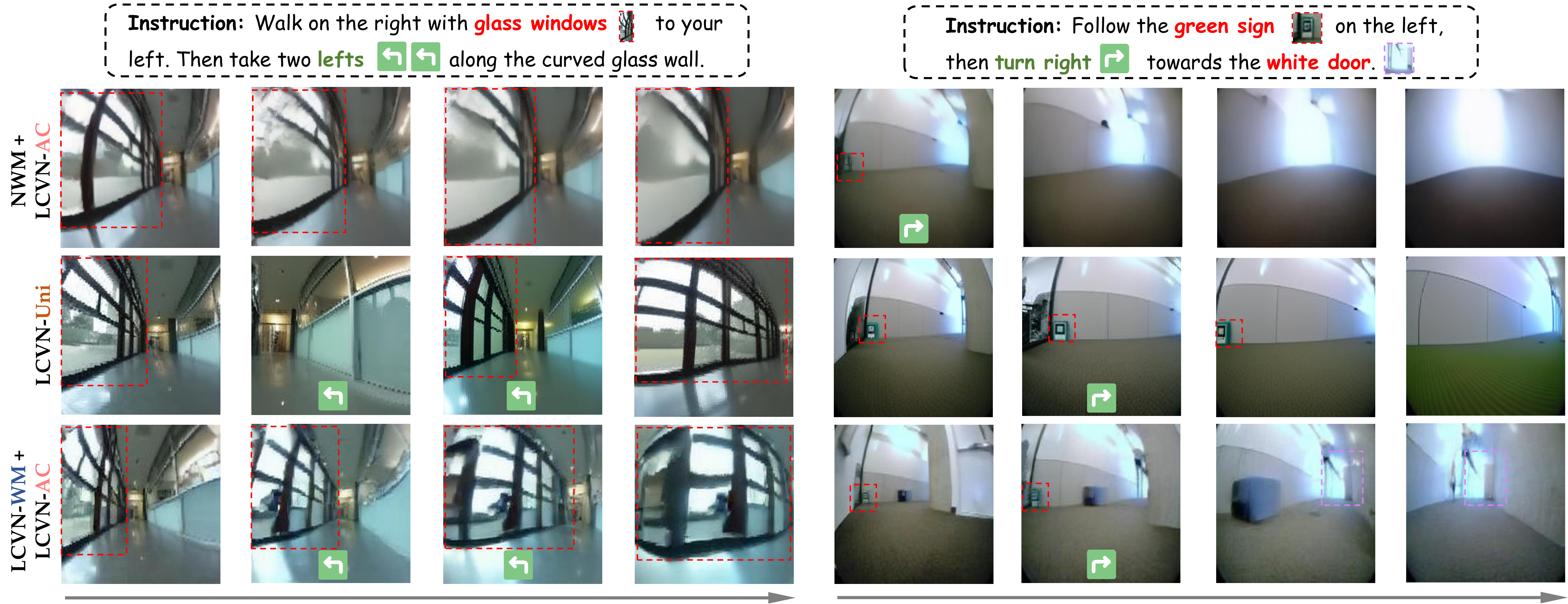}
    \caption{\small \textbf{Qualitative Comparisons} on LCVN val seen split across NWM and LCVN agents. LCVN agents exhibit stronger sensitivity to directional changes and are less prone to losing landmarks during world modeling.}
    \label{fig:qualitative-comp}

\end{figure*}

\subsection{Experimental Settings}
\label{sec:exp-settings}

\noindent \textbf{Baselines.}~We adopt three navigation baselines that leverage SOTA world models: Diamond~\citep{alonso2024diffusion} (UNet-based diffusion), NWM~\citep{bar2025navigation} (CDiT-based diffusion), and NWM (lang), our language-augmented variant of NWM where CLIP-encoded instructions are summed with action and time embeddings and fed through CDiT's AdaLN block. We further include two policy-only baselines, GNM~\citep{shah2022gnm} (lang) and NoMaD~\citep{sridhar2024nomad} (lang), adapted by replacing their image-goal encoder with a frozen CLIP text encoder followed by an MLP that projects instructions into the original goal-token space. Each baseline is retrained from scratch on the LCVN training split, with LCVN-AC trained specifically for each world-model baseline. To isolate language guidance, we evaluate two ablated variants: LCVN-Uni (w/o ins), which removes instructions only from observation-prediction substep, and LCVN-WM (w/o ins), trained without language conditioning.
\begin{table*}[!t]

\setlength{\abovecaptionskip}{2pt}
\setlength{\belowcaptionskip}{1pt}
    \centering 
        \caption{\small \textbf{Comparison with SOTA Methods} upon Language-Conditioned Visual Navigation on LCVN val seen, val unseen and test splits with SR, ATE, and RPE. We equip two LCVN agents with context sizes $k$ set to 1, 2, 4.}
    \resizebox{\linewidth}{!}{%
    \renewcommand\arraystretch{1.2}
        \small
    \begin{tabular}{lp{2cm}<{\centering}p{1cm}<{\centering}p{1cm}<{\centering}p{1cm}<{\centering}p{1cm}<{\centering}p{1cm}<{\centering}p{1cm}<{\centering}p{1cm}<{\centering}p{1cm}<{\centering}p{1cm}<{\centering}} 
    \hline\thickhline
        \multirow{2}{*}{\textbf{Methods}} & \multirow{2}{*}{\textbf{Context Size}} & \multicolumn{3}{c}{\textbf{Validation Seen}} & \multicolumn{3}{c}{\textbf{Validation Unseen}} & \multicolumn{3}{c}{\textbf{Test}} \\
        \cmidrule(lr){3-5} \cmidrule(lr){6-8} \cmidrule(lr){9-11}
        & & \textbf{ATE$\downarrow$} & \textbf{RPE$\downarrow$} & \textbf{SR$\uparrow$} & \textbf{ATE$\downarrow$} & \textbf{RPE$\downarrow$} & \textbf{SR$\uparrow$} & \textbf{ATE$\downarrow$} & \textbf{RPE$\downarrow$} & \textbf{SR$\uparrow$} \\
    \hline 
        GNM~\citep{shah2022gnm} (lang)  & 4 
        & 1.18 & 0.39 & 0.21  
        & 2.72 & 0.91 & 0.10  
        & 1.89 & 0.55 & 0.16 \\
        NoMaD~\citep{sridhar2024nomad} (lang)  & 4 
        & 1.08 & 0.36 & 0.24  
        & 2.61 & 0.88 & 0.11  
        & 1.78 & 0.53 & 0.18 \\
        Diamond~\citep{alonso2024diffusion} + LCVN-AC & 4 
        & 1.35 & 0.42 & 0.18  
        & 2.84 & 0.95 & 0.09  
        & 2.05 & 0.58 & 0.14 \\
        NWM~\citep{bar2025navigation} + LCVN-AC & 4 
        & 0.75 & 0.24 & 0.28  
        & 2.17 & 0.73 & 0.12  
        & 1.33 & 0.42 & 0.21 \\
        NWM~\citep{bar2025navigation} (lang) + LCVN-AC & 4 
        & 0.72 & 0.26 & 0.29  
        & 2.21 & 0.70 & 0.13  
        & 1.31 & 0.44 & 0.22 \\
    \hline 
        \rowcolor{w_1!3} LCVN-Uni & 1 
        & 0.43 & 0.13 & 0.37  
        & 1.36 & 0.52 & 0.21 
        & 0.74 & 0.25 & 0.32 \\
        \rowcolor{w_1!7} LCVN-Uni (w/o ins) & 2 
        & 0.47 & 0.15 & 0.34  
        & 1.49 & 0.58 & 0.19  
        & 0.81 & 0.28 & 0.29 \\
        \rowcolor{w_1!15} \textbf{LCVN-Uni} & \textbf{2} 
        & 0.36 & \textbf{0.11} & 0.42  
        & \textbf{1.21} & 0.47 & \textbf{0.27}  
        & \textbf{0.65} & 0.22 & \textbf{0.36} \\
        \rowcolor{w_1!21} LCVN-Uni & 4 
        & 0.37 & 0.12 & 0.41  
        & 1.24 & \textbf{0.43} & 0.25  
        & \textbf{0.65} & \textbf{0.21} & 0.35 \\
    \hline
        \rowcolor{w_1!3} LCVN-WM + LCVN-AC & 1 & 0.39 & 0.15 & 0.39 & 1.62 & 0.61 & 0.18 & 0.81 & 0.29 & 0.31 \\
        \rowcolor{w_1!7} LCVN-WM + LCVN-AC & 2 & 0.36 & 0.13 & 0.42 & 1.55 & 0.59 & 0.19 & 0.78 & 0.27 & 0.33 \\
        \rowcolor{w_1!15} LCVN-WM (w/o ins) + LCVN-AC & 4  & 0.46 & 0.16 & 0.35  
        & 1.82 & 0.67 & 0.15  
        & 0.89 & 0.31 & 0.29  \\
        \rowcolor{w_1!21} \textbf{LCVN-WM + LCVN-AC} & \textbf{4} & \textbf{0.34} & 0.12 & \textbf{0.43} & 1.51 & 0.56 & 0.19 & 0.76 & 0.26 & 0.34 \\

   \hline\thickhline 
    \end{tabular}}  
  \vspace{-0.2cm}
    \label{tab:planning_sota} 
\end{table*}

\begin{table*}[!t]
    \centering
        \caption{\small \textbf{Comparison with SOTA methods} on imagination performance (single-frame / long-horizon) on LCVN val seen split. We equip two LCVN agents with context sizes $k$ set to 1, 2, 4. Best results are in bold.}
    \renewcommand\arraystretch{1.4}
    \resizebox{\linewidth}{!}{%
    \begin{tabular}{lp{2.2cm}<{\centering}p{1.4cm}<{\centering}p{1.4cm}<{\centering}p{1.4cm}<{\centering}p{1.95cm}<{\centering}p{1.7cm}<{\centering}p{1.7cm}<{\centering}p{1.7cm}<{\centering}p{2.4cm}<{\centering}}
    \hline\thickhline
        \multirow{2}{*}{\textbf{Methods}} & \multirow{2}{*}{\textbf{Context Size}} & \multicolumn{4}{c}{\textbf{Single-frame Generation}} & \multicolumn{4}{c}{\textbf{Long-horizon Generation @8}} \\
        \cmidrule(lr){3-6} \cmidrule(lr){7-10}
        & & \textbf{SSIM $\uparrow$} & \textbf{PSNR $\uparrow$} & \textbf{LPIPS $\downarrow$} & \textbf{DreamSIM $\downarrow$} 
          & \textbf{SSIM$@8$ $\uparrow$} & \textbf{PSNR$@8$ $\uparrow$} & \textbf{LPIPS$@8$ $\downarrow$} & \textbf{DreamSIM$@8$ $\downarrow$} \\
    \hline
        Diamond~\citep{alonso2024diffusion} & 4 
        & 0.315 & 9.850 & 0.427 & 0.135 
        & 0.114 & 4.526  & 0.679 & 0.283 \\
        NWM~\citep{bar2025navigation} & 4 
        &0.370  &11.425  &0.314  & 0.096 
        &0.181  &6.057  & 0.629 & 0.196 \\
        NWM~\citep{bar2025navigation} (lang) & 4 
        & 0.382 & 11.612 & 0.309 & 0.094 
        & 0.176 & 6.041 & 0.605 & 0.191 \\
    \hline
        \rowcolor{w_1!3} LCVN-Uni & 1 
        &0.398 &12.881 &0.306 & 0.076
        &0.201 &7.057 &0.466 & 0.128\\
        \rowcolor{w_1!7} LCVN-Uni (w/o ins) & 2 
        &0.387 &12.642 &0.319 & 0.082
        &0.192 &6.874 &0.508 & 0.135\\
        \rowcolor{w_1!15} \textbf{LCVN-Uni} & \textbf{2} 
        &0.423 &13.325 &0.298 & \textbf{0.072}
        &0.218 &7.412 &0.451 & \textbf{0.119}\\
        \rowcolor{w_1!21} LCVN-Uni & 4 
        &0.421 &13.301 &0.293 & 0.073
        &0.216 &7.395 &0.445 & 0.121\\
    \hline
        \rowcolor{w_1!3} LCVN-WM & 1 
        &0.419 &18.895 &0.210 & 0.087
        &0.267 &10.530 &0.336 & 0.134\\
        \rowcolor{w_1!7} LCVN-WM & 2 
        &0.427 &19.742 &0.198 & 0.081
        &0.281 &11.084 &0.322 & 0.131\\
        \rowcolor{w_1!15} LCVN-WM (w/o ins) & 4 
        & 0.395 &17.247 & 0.236 & 0.118 
        & 0.241 &9.580 & 0.371 & 0.161 \\
        \rowcolor{w_1!21} \textbf{LCVN-WM} & \textbf{4} 
        &\textbf{0.435} &\textbf{20.316} &\textbf{0.189} & 0.078
        &\textbf{0.293} &\textbf{11.527} &\textbf{0.315} & 0.127\\
    \hline\thickhline
    \end{tabular}}
    \label{tab:frame_generation}
\end{table*}
\begin{table}[!t] 
\tabcolsep=0.18cm
\renewcommand\arraystretch{1.2}

     \centering 
          \caption{\small Ablations of \textbf{language, action and time conditioning} on navigation and imagination performance of LCVN-WM ($k$=4) on LCVN val seen split. \textcolor[rgb]{0,0,1}{\ding{51}} denotes with and \textcolor[rgb]{1,0,0}{\texttimes} denotes without. Best results are in bold.} 
     \resizebox{0.85\linewidth}{!}{%
     \begin{tabular}{p{1.5cm}<{\centering}p{1cm}<{\centering}p{1cm}<{\centering}p{0.8cm}<{\centering}p{0.8cm}<{\centering}p{0.8cm}<{\centering}p{1.1cm}<{\centering}p{2cm}<{\centering}p{2cm}<{\centering}p{2.5cm}<{\centering}} 
     \hline 
     &&&\multicolumn{3}{c}{\textbf{Navigation}}&\multicolumn{4}{c}{\textbf{Imagination}} \\ 
     \cmidrule(lr){4-6} \cmidrule(lr){7-10}       \textbf{Language}&\textbf{Action}&\textbf{Time}&\textbf{ATE}$\downarrow$&\textbf{RPE}$\downarrow$&\textbf{SR}$\uparrow$&\textbf{SSIM$\uparrow$} & \textbf{DreamSIM$\downarrow$} & \textbf{SSIM$@8$$\uparrow$} & \textbf{DreamSIM$@8$$\downarrow$}  \\\hline 
         \textcolor[rgb]{1,0,0}{\texttimes} & \textcolor[rgb]{1,0,0}{\texttimes} & \textcolor[rgb]{0,0,1}{\ding{51}} 
         & 1.82 & 0.63 & 0.12 
         & 0.251 & 0.627 
         & 0.095 & 0.798 \\ 
         
         \textcolor[rgb]{0,0,1}{\ding{51}} & \textcolor[rgb]{1,0,0}{\texttimes} & \textcolor[rgb]{1,0,0}{\texttimes} 
         & 1.12 & 0.35 & 0.19 
         & 0.341 & 0.137 
         & 0.187 & 0.226 \\ 
         
         \textcolor[rgb]{1,0,0}{\texttimes} & \textcolor[rgb]{0,0,1}{\ding{51}} & \textcolor[rgb]{1,0,0}{\texttimes} 
         & 0.54 & 0.22 & 0.31 
         & 0.388 & 0.112 
         & 0.229 & 0.169 \\

         \textcolor[rgb]{0,0,1}{\ding{51}} & \textcolor[rgb]{1,0,0}{\texttimes} & \textcolor[rgb]{0,0,1}{\ding{51}} 
         & 0.89 & 0.31 & 0.22 
         & 0.352 & 0.142 
         & 0.201 & 0.214 \\

          \textcolor[rgb]{0,0,1}{\ding{51}} & \textcolor[rgb]{0,0,1}{\ding{51}} & \textcolor[rgb]{1,0,0}{\texttimes} 
         & 0.37 & 0.14 & 0.41 
         & 0.422 & 0.081 
         & 0.275 & 0.138 \\ 
         
         \textcolor[rgb]{1,0,0}{\texttimes} & \textcolor[rgb]{0,0,1}{\ding{51}} & \textcolor[rgb]{0,0,1}{\ding{51}} 
         & 0.46 & 0.16 & 0.35 
         & 0.395 & 0.118 
         & 0.241 & 0.161 \\

        \rowcolor{w_1!21} \textcolor[rgb]{0,0,1}{\ding{51}} & \textcolor[rgb]{0,0,1}{\ding{51}} & \textcolor[rgb]{0,0,1}{\ding{51}} 
         & \textbf{0.34} & \textbf{0.12} & \textbf{0.43} 
         &\textbf{0.435} & \textbf{0.078} 
         &\textbf{0.293} & \textbf{0.127}\\ 
         \hline 
     \end{tabular}} 

     \label{tab:ablation_components} 
       \vspace{-0.2cm}
 \end{table}

\noindent \textbf{Evaluation Metrics.}~We evaluate performance using two suites of metrics.~\textbf{(1) Navigation quality.}~We report Absolute Trajectory Error (ATE), Relative Pose Error (RPE)~\citep{sturm2012evaluating}, and Success Rate (SR). A trajectory is considered successful if the final distance to goal is smaller than agent’s average step size (in meters).~\textbf{(2) Imagination quality.} To assess fidelity of navigation visualizations, we adopt structural and perceptual metrics including SSIM~\citep{wang2004ssim}, PSNR~\citep{hore2010image}, LPIPS~\citep{zhang2018unreasonable}, and DreamSim~\citep{fu2023dreamsim}. To further evaluate long-horizon stability under rollout, we report SSIM@n, PSNR@n, LPIPS@n, and DreamSim@n. Due to space limits, kindly refer to \textbf{Appendix} for details.

\noindent \textbf{Datasets \& Implementation Details.}~Experiments are conducted on our proposed LCVN dataset (Sec.~\ref{sec:LCVN-dataset}) using the val seen, val unseen, and test splits. For LCVN-WM and LCVN-AC, we train an ImageVAE~\citep{kingma2013auto} from scratch to compress images into $32 \times 32$ latents. LCVN-Uni is fine-tuned on GAIR Anole-7B~\citep{chern2024anole} (4096-token context), while keeping the text and image tokenizers as well as the bin-token encoder frozen. Input images are resized to $448 \times 448$ (height $\times$ width) and discretized into 784 visual tokens when context size is 1 or 2, and into 625 tokens when context size is 4. All models are trained on 4$\times$NVIDIA A100 GPUs (80GB each). Due to space limits, kindly refer to \textbf{Appendix} for additional implementation details.

\subsection{Comparisons to SOTA Methods}

\label{sec:sota-comp}
\noindent \textbf{Navigation performance.}~Table~\ref{tab:planning_sota} reveals a pattern aligned with structural differences between the two paradigms (qualitative results in Fig.~\ref{fig:qualitative-comp}). LCVN-WM~+~LCVN-AC achieves the strongest in-domain performance via high-fidelity LDiT rollouts, but this advantage shrinks on val unseen as rollouts may become unreliable on novel scenes. LCVN-Uni shows the opposite profile, leveraging Anole-7B's pre-trained multimodal prior and avoiding the two-stage error-compounding bottleneck. Absolute SR remains moderate across all methods, reflecting the inherent difficulty of LCVN: predicting entire trajectories without corrective feedback causes errors to compound over the horizon. Despite this, LCVN agents substantially surpass other baselines, establishing a strong starting point and validating both paradigms as promising directions.

\noindent \textbf{Imagination performance.}~Table~\ref{tab:frame_generation} compares imagination abilities of LCVN-Uni and LCVN-WM against SOTA baselines. Both LCVN agents achieve competitive results across single-frame and long-horizon evaluations. On one-step predictions, LCVN-WM ($k=4$) attains the highest structural fidelity, while LCVN-Uni ($k=2$) achieves the best perceptual alignment. Under multi-step rollouts, both agents maintain stability and mitigate compounding errors over longer horizons. 

\noindent \textbf{Failure Case Analysis.}~A representative failure case appears in Fig.~\ref{fig:qualitative-comp} (right): LCVN-Uni predicts the correct turning direction but only imagines a few post-turn view transitions: the corridor and target white door are not correctly synthesized. This reflects its weaker long-horizon imagination (Table~\ref{tab:frame_generation}), causing subsequent actions to lose visual grounding despite correct directional intent.

\subsection{Ablation Studies}

\begin{wrapfigure}{r}{0.46\textwidth}
\vspace{-0.3cm}
\setlength{\abovecaptionskip}{0pt} 
\setlength{\belowcaptionskip}{0pt} 
\centering
\includegraphics[width=0.46\textwidth]{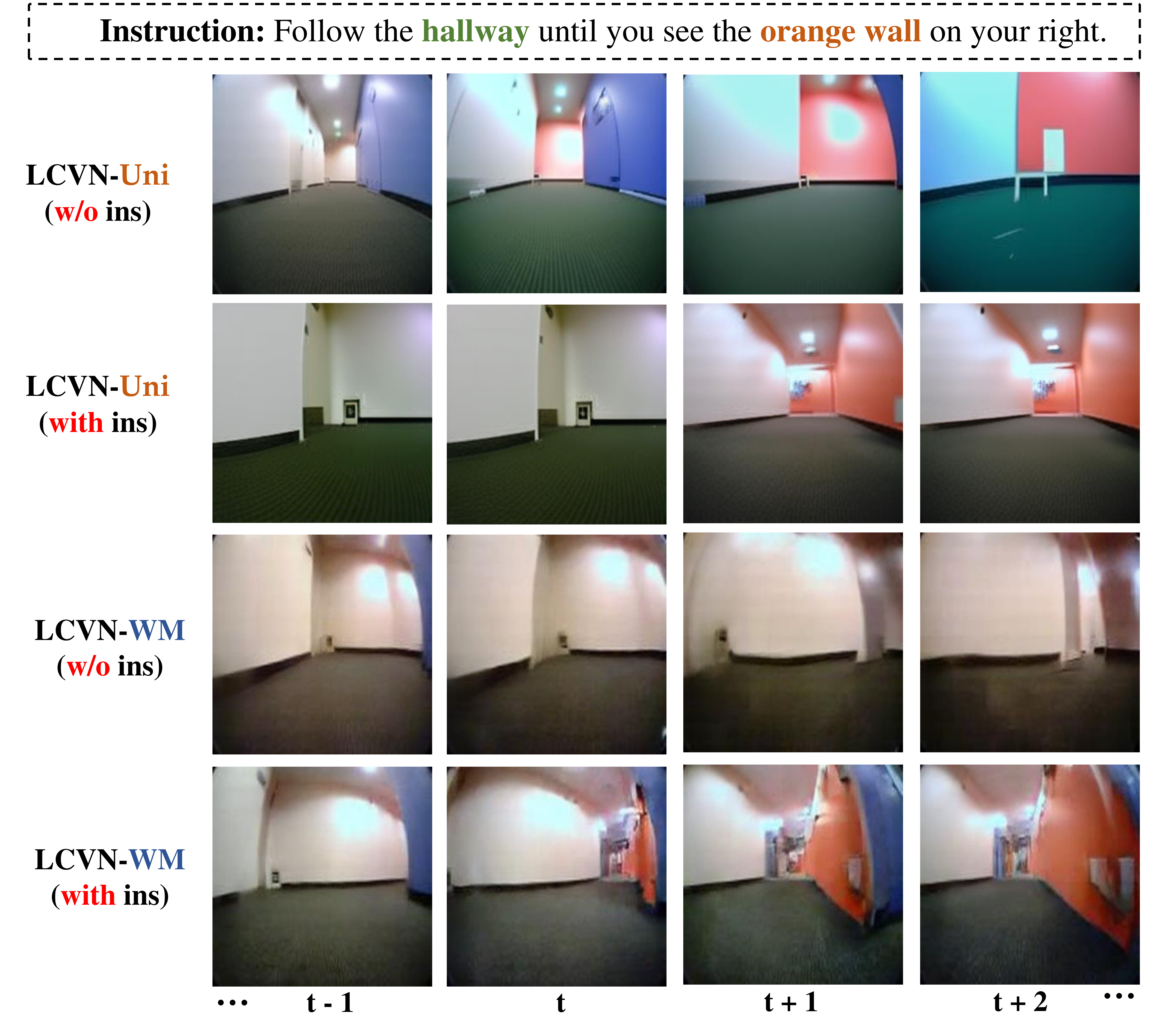}
\caption{\small Qualitative comparisons of language guidance showcase both baselines better preserve semantic consistency across states with language guidance.}
\label{fig:visual_result}
\vspace{-0.5cm}
\end{wrapfigure}

\noindent \textbf{1)~Language Guidance.}~We ablate the role of language guidance by comparing LCVN agents trained with and without instructions in Tables.~\ref{tab:planning_sota}, \ref{tab:frame_generation}, and Fig.~\ref{fig:visual_result}. Removing language consistently weakens both navigation and imagination: LCVN agents without instruction inputs exhibit lower success rates and less stable rollouts. In this task, language instructions serve as the goal specification, providing semantic alignment between states and destinations and thereby guiding effective navigation.

\noindent \textbf{2)~Context Size.}~Tables~\ref{tab:planning_sota} and~\ref{tab:frame_generation} show that larger contexts generally improve both LCVN methods. LCVN-Uni peaks at $k$~=~2 because Anole-7B's fixed 4096-token window forces a trade-off: images use 784 tokens at $k$~=~1,2 but only 625 at $k$~=~4, and the lower resolution offsets the longer horizon. We therefore adopt $k$~=~2 for LCVN-Uni and $k$~=~4 for LCVN-WM~+~LCVN-AC.

\begin{table}[!t] 

\tabcolsep=0.18cm
\renewcommand\arraystretch{1.2}
 \setlength{\abovecaptionskip}{0pt} 
 \setlength{\belowcaptionskip}{3pt} 
     \centering 
          \caption{\small Ablations of \textbf{instruction style} for LCVN-Uni ($k$=2) and LCVN-WM ($k$=4) on LCVN val seen split.} 
     \resizebox{0.8\linewidth}{!}{%
     \begin{tabular}{p{1.8cm}p{1.5cm}<{\centering}p{1.5cm}<{\centering}p{1cm}<{\centering}p{1cm}<{\centering}p{1.5cm}<{\centering}p{2.0cm}<{\centering}p{1.8cm}<{\centering}p{2.3cm}<{\centering}} 
     \hline 
     & & \multicolumn{3}{c}{\textbf{Navigation}} & \multicolumn{4}{c}{\textbf{Imagination}} \\ 
     \cmidrule(lr){3-5} \cmidrule(lr){6-9} 
     \textbf{Method} & \textbf{Ins. Style} 
     & \textbf{ATE$\downarrow$} & \textbf{RPE$\downarrow$} & \textbf{SR$\uparrow$} 
     & \textbf{SSIM$\uparrow$} & \textbf{DreamSIM$\downarrow$} 
     & \textbf{SSIM$@8$$\uparrow$} & \textbf{DreamSIM$@8$$\downarrow$} \\ 
     \hline 
     \multirow{3}{*}{\textbf{LCVN-Uni}} & Concise   & 0.37 & 0.13 & 0.41 & 0.421 & 0.076 & 0.218 & 0.121 \\ 
     & Intricate & 0.39 & 0.11 & 0.38 & 0.415 & 0.074 & 0.211 & 0.123 \\ 
     \rowcolor{w_1!21} \cellcolor{white!13}& \textbf{Landmark.} 
       & \textbf{0.32} & \textbf{0.10} & \textbf{0.47} 
       & \textbf{0.433} & \textbf{0.069} & \textbf{0.225} & \textbf{0.116} \\ 
     \hline 
     & Concise   & 0.35 & 0.12 & 0.42 & 0.435 & 0.078 & 0.293 & 0.127 \\ 
     & Intricate & 0.36 & 0.14 & 0.40 & 0.429 & 0.081 & 0.285 & 0.128 \\ 
     \rowcolor{w_1!21} \cellcolor{white!13}\multirow{-3}{*}{\textbf{LCVN-WM}} & \textbf{Landmark.} 
       & \textbf{0.31} & \textbf{0.11} & \textbf{0.47} 
       & \textbf{0.442} & \textbf{0.076} & \textbf{0.302} & \textbf{0.125} \\ 
     \hline 
     \end{tabular}} 
       \vspace{-0.2cm}
     \label{tab:instruction_style} 
 \end{table}
 \begin{table}[!t] 
 \renewcommand\arraystretch{1.2}
 \tabcolsep=0.18cm
 \setlength{\abovecaptionskip}{2pt} 
 \setlength{\belowcaptionskip}{1pt} 
     \centering 
          \caption{\small \textbf{Pixel vs. Latent space} on the performance of LCVN-WM ($k$=4) on LCVN val seen split.}
 
     \resizebox{0.8\linewidth}{!}{%
     \begin{tabular}{p{2.8cm}p{1.2cm}<{\centering}p{1.2cm}<{\centering}p{1.2cm}<{\centering}p{1.2cm}<{\centering}p{2cm}<{\centering}p{1.7cm}<{\centering}p{2.5cm}<{\centering}} 
     \hline 
      & \multicolumn{3}{c}{\textbf{Navigation}} & \multicolumn{4}{c}{\textbf{Imagination}} \\ 
     \cmidrule(lr){2-4} \cmidrule(lr){5-8} 
     \textbf{Encoding Space} 
     & \textbf{ATE$\downarrow$} & \textbf{RPE$\downarrow$} & \textbf{SR$\uparrow$} 
     & \textbf{SSIM$\uparrow$} & \textbf{DreamSIM$\downarrow$} 
     & \textbf{SSIM$@8$$\uparrow$} & \textbf{DreamSIM$@8$$\downarrow$} \\ 
     \hline 
     Pixel 
     & 0.42 & 0.16 & 0.36 
     & 0.381 & 0.095 
     & 0.257 & 0.150 \\ 
     \rowcolor{w_1!21} \textbf{Latent} 
     & \textbf{0.34} & \textbf{0.12} & \textbf{0.43} 
     & \textbf{0.435} & \textbf{0.078} 
     & \textbf{0.293} & \textbf{0.127}\\ 
     \hline 
     \end{tabular}} 
 
     \label{tab:encoding_comparison} 
       \vspace{-0.2cm}
 \end{table}
\noindent \textbf{3)~Language, Action and Time Conditioning.}~We train LCVN-WM with joint conditioning on language, action, and time, and ablate each in Table~\ref{tab:ablation_components}. Time conditioning alone is insufficient for meaningful navigation. Adding action conditioning on top of time yields a substantial improvement,
as explicit control signals are essential for grounding latent dynamics.

Further adding language conditioning produces an additional gain, with language providing semantic alignment between states and goals to disambiguate where action-driven dynamics should lead. This progression indicates that primary bottleneck lies in grounding linguistic intent into continuous control.


\noindent\textbf{4)~Instruction Style.}~Table~\ref{tab:instruction_style} shows that landmark-grounded instructions consistently yield the best performance for both LCVN baselines, while concise instructions remain better than intricate ones. Intricate descriptions add scene detail not strictly relevant to navigation, indicating that navigation-relevant signal density rather than descriptive richness drives instruction-following quality.

\noindent\textbf{5)~Latent \emph{vs.} Pixel Space.}~Table~\ref{tab:encoding_comparison} shows that training LCVN-WM in pixel space (PatchEmbed~\citep{dosovitskiy2020image} following~\citep{song2025history}) consistently underperforms its latent counterpart. Latent encoding also provides a compact shared representation that integrates naturally with LCVN-AC under Paradigm 1.

\noindent \textbf{6)~External Data Scaling.}~Following the original NWM protocol, we train both NWM and LCVN-WM with additional Ego4D~\citep{grauman2022ego4d} data. Table~\ref{tab:external_data} confirms that external data benefits generalization, with LCVN-WM maintaining a consistent advantage over NWM under both settings.

\begin{table}[!t]
\renewcommand\arraystretch{1.2}
\tabcolsep=0.18cm
    \centering
        \caption{\small \textbf{External data scaling} on imagination performance of NWM and LCVN-WM ($k$=4) on LCVN val unseen. \textcolor[rgb]{0,0,1}{\ding{51}} denotes training with additional Ego4D data. Best results are in bold.}
    \resizebox{0.8\linewidth}{!}{%
    \begin{tabular}{p{2cm}p{2.3cm}<{\centering}p{1.7cm}<{\centering}p{2.2cm}<{\centering}p{2cm}<{\centering}p{2.5cm}<{\centering}}
    \hline
    \textbf{Method} & \textbf{External Data} 
    & \textbf{SSIM$\uparrow$} & \textbf{DreamSIM$\downarrow$} 
    & \textbf{SSIM$@8$$\uparrow$} & \textbf{DreamSIM$@8$$\downarrow$} \\
    \hline
        NWM & \textcolor[rgb]{1,0,0}{\texttimes} 
        & 0.180 & 0.325 & 0.102 & 0.543 \\
        NWM & \textcolor[rgb]{0,0,1}{\ding{51}} (Ego4D) 
        & 0.197 & 0.318 & 0.115 & 0.528 \\
    \hline
        LCVN-WM & \textcolor[rgb]{1,0,0}{\texttimes} 
        & 0.302 & 0.239 & 0.167 & 0.376 \\
        \rowcolor{w_1!21} \textbf{LCVN-WM} & \textcolor[rgb]{0,0,1}{\ding{51}} \textbf{(Ego4D)} 
        & \textbf{0.313} & \textbf{0.226} & \textbf{0.184} & \textbf{0.357} \\
    \hline
    \end{tabular}}
    \label{tab:external_data}
\end{table}
\section{Conclusion}
\label{sec:conclusion}

We introduced language-conditioned visual navigation (LCVN), a natural language variant of goal-conditioned visual navigation, supported by a large-scale LCVN dataset, and instantiated two paradigms: latent-imagination policy learning (LCVN-WM~+~LCVN-AC), which imagines futures with a diffusion world model and learns a latent-space actor--critic, and unified autoregressive prediction (LCVN-Uni), which jointly predicts actions and observations. Experiments show that LCVN-WM~+~LCVN-AC produces more temporally coherent rollouts while LCVN-Uni generalizes better to unseen environments, and ablations further surface three findings: language guidance acts as the goal specification that aligns states with destinations and stabilizes long-horizon rollouts; the primary bottleneck lies in grounding linguistic intent into continuous control; and navigation-relevant signal density matters more than descriptive richness. We hope LCVN serves as a testbed for studying how language, imagination, and decision-making interact in embodied agents.

\section*{Acknowledgments}
This work was supported in part by faculty startup funding from the University of Washington and by the U.S. Department of Transportation through the Pacific Northwest Transportation Consortium (PacTrans), including its seed funding and small research projects programs. The authors also acknowledge the Vicky L. Carwein and William B. Andrews Endowment for Graduate Programs and the University of Washington Graduate School Top Scholar Awards. The authors further acknowledge the Undergraduate Research Faculty Fellows program at the University of Washington, supported by the Henry Luce Foundation, and the professional development support provided through Engineering Unleashed and the Kern Entrepreneurial Engineering Network (KEEN), supported by The Kern Family Foundation.

\appendix
\section*{Technical appendices and supplementary material}

\setcounter{table}{0}  
\setcounter{figure}{0}

\setcounter{equation}{0}
\renewcommand{\thetable}{A\arabic{table}}
\renewcommand{\thefigure}{A\arabic{figure}}

\renewcommand{\theequation}{A\arabic{equation}}

This supplementary material provides additional details and results that complement the main paper. Sec.~\ref{appx.LCVN-dataset} presents multi‑style data samples from the LCVN dataset. Sec.~\ref{sec:appx-method-details} includes pseudo‑code for the LCVN‑WM and LCVN‑AC, as well as prompt and training details for the LCVN‑Uni. Sec.~\ref{sec:appx-exp-details} reports further experimental details and more experiments.

\section{LCVN Dataset Details}
\label{appx.LCVN-dataset}

\begin{figure*}[t]
    \centering    
   \includegraphics[width=0.999\linewidth]{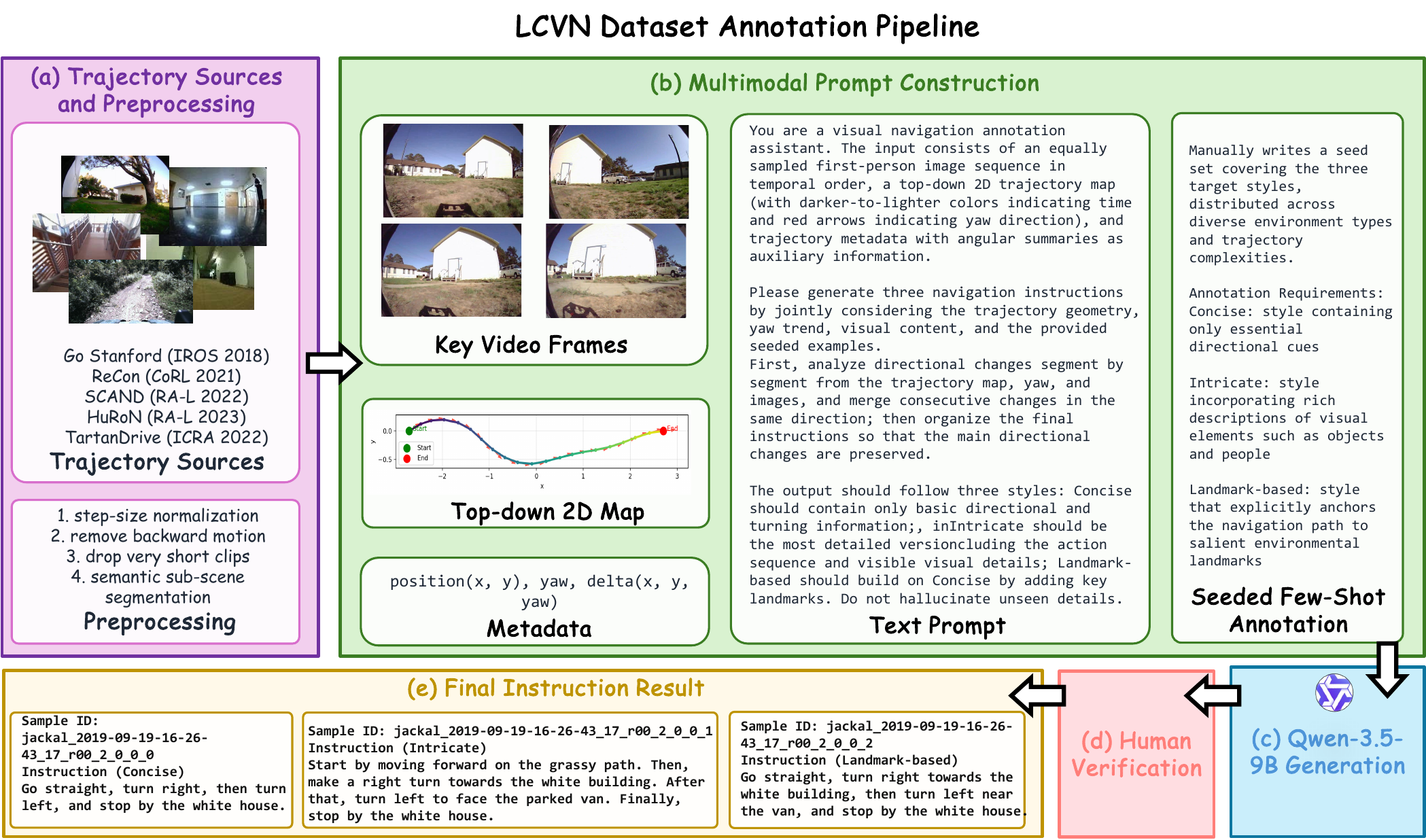}
    \caption{\small Overview of \textbf{annotation pipeline} of the LCVN dataset.}
    \label{fig:lcvn-dataset-pipeline}
\end{figure*}

As shown in Fig.~\ref{fig:lcvn-dataset-pipeline}, this section complements Sec.~\ref{sec:LCVN-dataset} with details on the trajectory sources, the seeded few-shot annotation pipeline, human evaluation, and qualitative samples of the three instruction styles.

\subsection{Trajectory Sources and Preprocessing}

Go Stanford, ReCon, SCAND, and HuRoN provide diverse indoor and structured outdoor trajectories that populate the training and val seen splits, while TartanDrive contributes exclusively to val unseen and partially to test. TartanDrive's off-road driving scenes differ markedly from the other four sources in terrain, camera dynamics, and visual appearance, providing a strong distribution shift for zero-shot evaluation. The test split intentionally blends held-out trajectories from all five sources so that a single evaluation simultaneously measures in-domain proficiency and out-of-domain generalization. To standardize trajectories from heterogeneous sources, we normalize per-frame displacement by each source's average step size to yield an embodiment-agnostic continuous control space, filter out backward movements and trajectories shorter than three steps following~\citep{bar2025navigation, sridhar2024nomad}, and segment each visual stream into semantically coherent sub-scenes using Qwen-3.5-9B~\citep{qwen35blog}. 

\subsection{Seeded Few-Shot Annotation Pipeline}

Our annotation team first manually writes a seed set covering the three target styles, distributed across diverse environment types and trajectory complexities (straight, single-turn, multi-turn). Annotators are instructed to write naturally and idiomatically, with attention to stylistic markers that distinguish the three styles: terse imperatives for Concise, descriptive subordinate clauses for Intricate, and landmark-anchored prepositional phrases for Landmark-based. This seed set serves as the linguistic anchor for all subsequent generation.

For each target trajectory, we render a top-down 2D map alongside the egocentric video. The map plots the agent's planar trajectory using local odometry, with the start point marked, the heading direction indicated by an arrow at each waypoint, and turning angles annotated; sampled camera frustums convey the egocentric viewpoint orientation. Without this 2D map as visual input, the VLM frequently misinterprets terrain-induced camera shake or local sidesteps as deliberate global heading changes, producing instructions like ``turn right'' for trajectories that are globally straight. The 2D map disambiguates global heading from local motion noise, substantially improving geometric correctness of generated instructions. 

We then construct a style-specific few-shot prompt for each trajectory by sampling seeds matched to the target style, environment type, and complexity, and pass this prompt along with the target trajectory's egocentric frames and 2D map to Qwen-3.5-9B. We generate three candidate instructions per trajectory (one per style). All generated instructions are subsequently reviewed by our annotation team, who cross-reference them with the corresponding video and 2D map and check for geometric correctness, landmark accuracy, style adherence, and naturalness. Annotators correct geometric or landmark errors and rewrite instructions that drift from the target style. The final dataset thus combines human authorship at the seed stage, VLM scaling at the generation stage, and human verification at the refinement stage, rather than the simple ``machine-generated then verified'' pipeline that motivates concerns about linguistic homogeneity.

\subsection{Human Quality Evaluation}

To quantitatively assess the quality of the final instructions, we conduct a human evaluation on a random subset of the dataset. We sample 640 trajectories roughly balanced across the five sources and three instruction styles, and recruit volunteer evaluators from our research group who are not involved in the annotation pipeline. Each evaluator watches the egocentric video, views the corresponding 2D trajectory map, and rates the paired instruction on five dimensions using a 1--6 Likert scale, where 6 indicates the instruction is indistinguishable from expert human-written ones: (1)~\emph{Trajectory--Instruction Alignment}, whether the instruction accurately describes the path including turns, distances, and stopping point; (2)~\emph{Style Adherence}, whether the instruction conforms to its target style without drifting into characteristics of another; (3)~\emph{Geometric Correctness}, whether directional references and spatial relations are consistent with egocentric viewpoint; (4)~\emph{Landmark Accuracy}, whether mentioned visual landmarks are present and correctly localized; and (5)~\emph{Naturalness \& Fluency}, whether the instruction reads as natural language a human guide would produce. Each trajectory--instruction pair is rated by three evaluators, and scores are averaged across evaluators and dimensions.

The overall mean score across all 640 samples is 5.2 out of 6.0, indicating that the vast majority of instructions are rated close to expert human quality. These results confirm that the combination of seeded few-shot generation, 2D map conditioning, and human verification yields instructions of consistently high quality across all three styles.

\subsection{Multi-Style Instruction Examples}

Table~\ref{appx.tab:LCVN-dataset-example} presents two representative trajectories, each annotated in all three styles, with text in \textcolor{purple}{purple} marking directional guidance and text in \textcolor{blue}{blue} marking landmark references. The Concise instructions provide minimal yet sufficient directional guidance, the Intricate instructions enrich the path with descriptions of surrounding objects and scene context, and the Landmark-based instructions explicitly tie actions to recognizable environmental features. The three styles are designed to be complementary rather than redundant: Concise probes whether models can follow minimal directional language, Intricate probes whether models can extract navigationally-relevant content from verbose descriptions, and Landmark-based probes whether models can ground language into specific visual features. Comparing model performance across styles in Table~\ref{tab:instruction_style} thus serves as a structured diagnostic of where language grounding succeeds or fails.

\begin{table*}[!t]
 \setlength{\abovecaptionskip}{3pt} 
 \setlength{\belowcaptionskip}{1pt}
\centering
\caption{\small \textbf{Multi-style Instruction Samples from the LCVN Dataset.} Text in \textcolor{purple}{purple} denotes directional guidance, while text in \textcolor{blue}{blue} marks landmark references. Each trajectory in LCVN is paired with three distinct instruction styles (concise, intricate and landmark-grounded), enabling a comprehensive evaluation of model generalization across varying levels of specificity and emphasis~\citep{kolagar2024aligning}.}
\resizebox{\linewidth}{!}{%
\begin{tabular}{|p{1\textwidth}|}
\hline
\textbf{Trajectory example 1 (concise style).} 

\textcolor{purple}{Walk straight} ahead and \textcolor{purple}{stop} at the intersection.
 \\
\hline
\textbf{Trajectory example 1 (intricate style). }

You are on a wide sidewalk lined with \textcolor{blue}{trees} on both sides. A \textcolor{blue}{pedestrian} is approaching from the front. \textcolor{purple}{Continue walking} until a \textcolor{blue}{red stall} appears on your right, with \textcolor{blue}{three people} gathered around it, and \textcolor{purple}{stop} in front of the \textcolor{blue}{stall}.
 \\
\hline
\textbf{Trajectory example 1 (landmark-grounded style).} 

\textcolor{purple}{Walk along} the sidewalk until you reach a \textcolor{blue}{red stall} ahead, and \textcolor{purple}{stop} in front of it.
 \\
\hline

\textbf{Trajectory example 2 (concise style).} 

\textcolor{purple}{Walk straight} ahead, \textcolor{purple}{turn left} at the first corner, and \textcolor{purple}{stop} along the wall.
 \\
\hline
\textbf{Trajectory example 2 (intricate style). }

\textcolor{purple}{Proceed straight} along the gray corridor, passing a row of \textcolor{blue}{white cabinets} and a \textcolor{blue}{glass-walled office} on your left where people appear to be working. \textcolor{purple}{Continue forward} until you reach a \textcolor{blue}{gray door}, with a \textcolor{blue}{small white trash bin} positioned to its left. Then \textcolor{purple}{turn left} into another gray corridor and \textcolor{purple}{stop} along the wall on your left.
 \\
\hline
\textbf{Trajectory example 2 (landmark-grounded style).} 

\textcolor{purple}{Walk straight} down the corridor, passing a \textcolor{blue}{glass-walled office} on your left. Continue until you reach a \textcolor{blue}{gray door}, then \textcolor{purple}{turn left} and \textcolor{purple}{stop} along the wall on your left.
 \\

\hline
\end{tabular}}
\label{appx.tab:LCVN-dataset-example}

\end{table*}

\section{Additional Implementation Details}
\label{sec:appx-method-details}

\subsection{Pseudo-Code for LCVN-WM \& LCVN-AC}
\label{sec:appx-pseudocode}

Fig.~2 in the main text illustrates the overall architecture of both LCVN-WM and LCVN-AC, highlighting three stages of training and inference. To complement this high-level overview, we provide detailed pseudo-code here (Algorithms~\ref{alg:LCVN_wm}, \ref{alg:LCVN_ac} and \ref{alg:LCVN_inference}) that specifies the step-by-step procedures underlying each stage.

\noindent \textbf{Training Phase 1 (LCVN-WM).} Algorithm~\ref{alg:LCVN_wm} outlines the training procedure of the LCVN-WM. LCVN-WM incorporates language conditioning by extending a diffusion transformer with cross-attention to CLIP-encoded instructions, together with action and time-shift conditioning. In addition, it adopts Diffusion Forcing, which applies heterogeneous noise levels across latent context window. This mechanism strengthens temporal modeling and enables the model to predict future latent states conditioned on actions and instructions.

\begin{algorithm}[t]
\caption{Training Phase 1: LCVN-WM with diffusion forcing}
\label{alg:LCVN_wm}
\DontPrintSemicolon
\KwIn{Trajectory $\text{observations }o_{0:n}$, $\text{actions } a_{0:n-1}$, $\text{instruction } I$, $\text{timeshift } t_s$, VAE encoder Enc, language encoder CLIPTextEnc, MLP layer MLP, world model $F_\theta$, context size $k$}
\While{not converged}{
  Sample a trajectory $(o_{0:n}, a_{0:n-1}, I, t_s)$\;
  \For{$i \leftarrow 0$ \KwTo $n$}{
    $s_i \leftarrow \text{Enc}(o_i)$ \tcp*{encode images to latents}
  }
  $I_{\text{clip}} \leftarrow \text{CLIPTextEnc}(I)$ 
  $\mathcal{L}_{\text{WM}} \leftarrow 0$\;
  \For{$t \leftarrow k-1$ \KwTo $n-1$}{ 
    Build context window $\mathbf{s}_t \leftarrow [s_t, s_{t-1}, \dots, s_{t-k+1}]$\;
    \tcp{Diffusion Forcing}
    \For{$m \leftarrow 0$ \KwTo $k-1$}{
      Sample noise level $\ell_m$ and $\epsilon_m \sim \mathcal{N}(0, I)$\;
      $\tilde{s}_{t-m} \leftarrow \sqrt{\alpha_{\ell_m}}\, s_{t-m}
        + \sqrt{1-\alpha_{\ell_m}}\, \epsilon_m$ (Eq.~2)\;
    }
    $\tilde{\mathbf{s}}_t \leftarrow [\tilde{s}_t, \tilde{s}_{t-1}, \dots, \tilde{s}_{t-k+1}]$\;
    $e_a \leftarrow \text{MLP}(a_t)$;
    $e_{ts} \leftarrow \text{MLP}(t_s)$\; $e_d \leftarrow \text{MLP}(t_{\text{diff}})$\;
    $e_{\text{cond}} \leftarrow e_a + e_{ts} + e_d$\;
    $\hat{s}_{t+1} \leftarrow F_\theta(\tilde{\mathbf{s}}_t, I_{\text{clip}}, e_{\text{cond}})$\;
    $\mathcal{L}_{\text{WM}} \mathrel{+}= 
      \mathcal{L}_{\text{diff}}(\hat{s}_{t+1}, s_{t+1}, t_{\text{diff}})$\;
  }
    Update $\theta$\;
}
\end{algorithm}

\noindent \textbf{Training Phase 2 (LCVN-AC).} Algorithm~\ref{alg:LCVN_ac} details the actor–critic agent trained entirely in the latent space of LCVN-WM. The learner encoder produces latent plans conditioned on current states and instructions, which are aligned with expert plans via KL divergence. Actor–critic optimization incorporates intrinsic rewards that measure agreement between predicted and expert rollouts, while the critic regresses to $\lambda$-returns. The actor is regularized by both plan consistency and language alignment losses, ensuring semantically grounded navigation behavior.

\noindent \textbf{Inference Stage.} Algorithm~\ref{alg:LCVN_inference} outlines the coupled inference procedure. Given an initial observation and instruction, LCVN-WM imagines future latent states, while LCVN-AC generates corresponding actions conditioned on both the imagined states and instruction embeddings. The agent executes actions until a stop command is issued, thereby integrating language grounding, imagination, and policy learning into a unified navigation loop.

Together, these pseudo-code listings complement the schematic in Fig.~2, clarifying how LCVN-WM and LCVN-AC interact across training and inference to realize language-conditioned visual navigation.

\begin{algorithm}[t]
\caption{Training Phase 2: LCVN-AC with Intrinsic Rewards}
\label{alg:LCVN_ac}
\DontPrintSemicolon
\KwIn{Frozen LCVN-WM $F_\theta$, latent observations $s^E_{0:n}$, actions $a_{0:n-1}$, instruction $I$, timeshift $t_s$, context size $k$, horizon $H$}
\KwIn{Expert encoder $q_{\text{expert}}$, learner encoder $q_{\text{learner}}$, actor $\pi_\theta$, critic $v_\psi$}
\While{not converged}{
  Sample a trajectory $(s^E_{0:n}, a_{0:n-1}, I)$\;
  $I_{\text{clip}} \leftarrow \text{CLIPTextEnc}(I)$\;
  Initialize losses: $\mathcal{L}_{\text{critic}} \leftarrow 0,\ \mathcal{L}_{\text{actor}} \leftarrow 0$\;
  \For{$t \leftarrow 0$ \KwTo $n-1$}{
    Expert plan: $z_{\text{expert}} \sim q_{\text{expert}}(z \mid s^E_{t:n}, I_{\text{clip}})$\;
    Learner plan: $z_{\text{plan}} \sim q_{\text{learner}}(z \mid s^E_t, I_{\text{clip}})$\;
    Compute KL divergence (Eq.~3): 
    $\mathcal{L}_{\text{KL}}^t \leftarrow D_{\text{KL}}(q_{\text{expert}} \,\|\, q_{\text{learner}})$\;
    Build latent history $\mathbf{s}_t \leftarrow [s^E_t, s^E_{t-1}, \dots, s^E_{t-k}]$ (pad if $t < k$)\;
    \For{$k \leftarrow t$ \KwTo $\min(t+H-1, n-1)$}{
      Sample action: $\hat{a}_k \sim \pi_\theta(s^E_k, I_{\text{clip}}, z_{\text{plan}})$\;
     Predict next latent with WM (Eq.~4):\; \Indp $\hat{s}^\pi_{k+1} \sim F_\theta(\hat{s}^\pi_{k+1} \mid \mathbf{s}_k, \hat{a}_k, I_{\text{clip}}, t_s)$\; \Indm
      Compute intrinsic reward $r^{\text{int}}_k$ (Eq.~5)\;
      Append $r^{\text{int}}_k$, $\hat{s}^\pi_{k+1}$; update context $\mathbf{s}_{k+1}$\;
    }
    Compute discounted returns (Eq.~6)\;
    \For{each rollout step $k$}{
      Compute critic loss $\mathcal{L}_{\text{critic}}$ (Eq.~7)\;
      Compute actor loss $\mathcal{L}_{\text{actor}}$ (Eq.~8)\;
    }
  }
  Update parameters $\psi$, $\theta$, $q_{\text{learner}}$\;
}
\end{algorithm}

\begin{algorithm}[t]
\caption{Inference with coupling LCVN-WM and LCVN-AC}
\label{alg:LCVN_inference}
\DontPrintSemicolon
\KwIn{VAE encoder Enc, LCVN-WM $F_\theta$, learner encoder $q_{\mathrm{learner}}$, actor $\pi_\phi$}
\KwIn{Instruction $I$, initial observation $o_s$, context size $k$, max steps $T_{\max}$}
Encode instruction: $I_{\mathrm{clip}} \leftarrow \mathrm{CLIPTextEnc}(I)$\;
Encode initial observation: $s_0 \leftarrow \mathrm{Enc}(o_s)$\;
Initialize context $\mathbf{s}$ (length $k$) with $s_0$ \;
Initialize previous action $a_{\mathrm{prev}} \leftarrow \mathbf{0}$\;
\For{$t \leftarrow 0$ \KwTo $T_{\max}-1$}{
  \If{$t > 0$}{
    \tcp{imagine the next visual latent state}
    $\hat{s}_t \sim F_\theta(\hat{s}_t \mid \mathbf{s}, a_{\mathrm{prev}}, I_{\mathrm{clip}}, t_s)$\;
    Update $\mathbf{s}$ by appending $\hat{s}_t$ and removing the oldest latent\;
  }
  Sample plan: $z_{\mathrm{plan}} \sim q_{\mathrm{learner}}(z \mid s_t, I_{\mathrm{clip}})$\;
  \tcp{predict the next action}
  $\hat{a}_t \sim \pi_\phi(\cdot \mid s_t, I_{\mathrm{clip}}, z_{\mathrm{plan}})$\;
  \If{$\hat{a}_t = \mathbf{0}$}{
    \textbf{break} \tcp*{stop action to terminate the process}
  }
  $a_{\mathrm{prev}} \leftarrow \hat{a}_t$\;
}
\Return Executed trajectory $\{\hat{a}_0, \dots, \hat{a}_T\}$\;
\end{algorithm}

\subsection{Prompt Design for LCVN-Uni}
\label{appx. prompt}

\begin{figure*}[!t]
\newtcolorbox{fullpromptbox}[3]{
  enhanced,
  colback=#1!5,
  colframe=#1!50,
  fonttitle=\bfseries,
  title=#2,
  sharp corners,
  boxrule=0.8pt,
  width=\linewidth,
  lower separated=false,
  bottomtitle=0mm
}
\begin{fullpromptbox}{teal}{LCVN-Uni Prompt}
\textbf{\tcbox[colback=gray!20, colframe=gray!20, boxrule=0.1pt,top=0mm, bottom=0mm]{\quad\quad\quad\quad\quad\quad\quad\quad\quad\quad\quad\quad\quad\quad\quad\quad\quad\quad\quad\quad \textbf{Input}\quad\quad\quad\quad\quad\quad\quad\quad\quad\quad\quad\quad\quad\quad\quad\quad\quad\quad\quad\quad}}

\textbf{Task:} Joint Navigation Single Step Planning and Visualization

\textbf{Description:} Based on a given navigation instruction, the current first-person observation, and starting point observation, predict the next action to take. Also, predict the next first-person view observation after the agent executes this navigation action.

\textbf{Inputs:} Current Action: Move by dx: -0.16, dy: 0.2, dyaw: -0.02

Navigation Instruction: Move across the grass toward the white house on the front‑left, and stop in front of it.



\begin{tabular}{cc}
\centering
\quad\quad\quad\quad\quad\quad~ ~~~~ ~ ~~~Start Observation:~~~~~~ ~ ~ ~  & ~ ~ ~~~~Current Observation:~~~ ~ ~ ~ \\
\quad\quad\quad\quad\quad\quad\quad\quad\includegraphics[height=1.7cm,width=2.265cm]{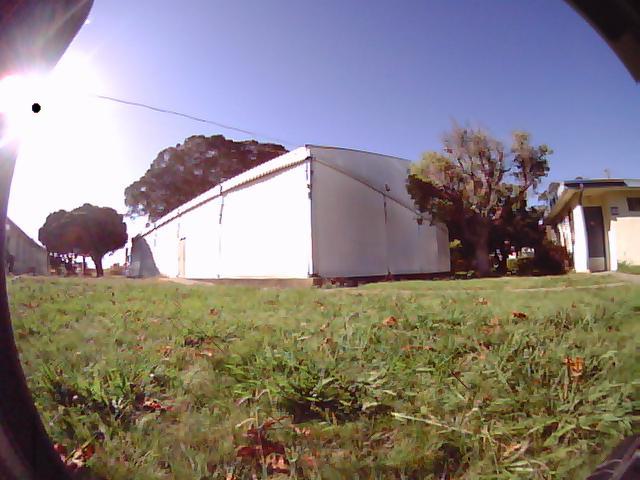}\quad\quad\quad &
\includegraphics[height=1.7cm,width=2.265cm]{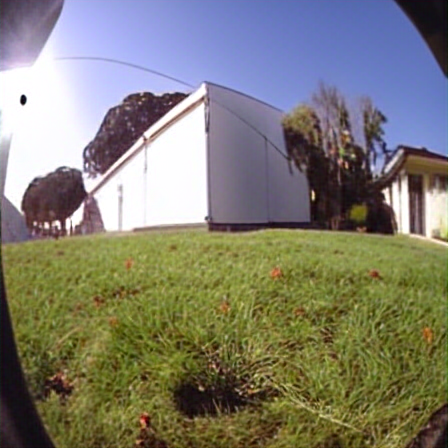}
\end{tabular}

\textbf{Action Format:}~The action can be the language command 'Stop', indicating the end of the trajectory. Alternatively, the action can be shifts composed of three components: - dx: displacement along the agent's facing direction), - dy: displacement perpendicular to the facing direction), - dyaw: change in heading angle (i.e., how much the agent rotates). All components are discretized into bin tokens: for example, - `dx pos bin 02`: dx = +0.02 meters, - `dy neg bin 23`: dy = -0.23 meters, - `dyaw pos bin 26`: counterclockwise rotation of +0.26 radians.

\textbf{Spatial Interpretation:} The magnitude of [dx, dy] reflects how far the agent moves in this step — larger values indicate greater positional shift, leading to larger visual changes. - dyaw controls the agent's rotation. A positive dyaw indicates a left turn (counter-clockwise), while a negative dyaw indicates a right turn (clockwise).

\textbf{Goal:}~Predict the most likely next action and next first-person observation, considering how the movement and rotation implied by `dx`, `dy`, and `dyaw` would affect what the agent sees next.

\textbf{\tcbox[colback=gray!20, colframe=gray!20, boxrule=0.1pt,top=0mm, bottom=0mm]{\quad\quad\quad\quad\quad\quad\quad\quad\quad\quad\quad\quad\quad\quad\quad\quad\quad\quad\quad\quad\quad\quad \textbf{Response} \quad\quad\quad\quad\quad\quad\quad\quad\quad\quad\quad\quad\quad\quad\quad\quad\quad\quad\quad\quad\quad}}
\textbf{Predicted action:} Move by dx:~$<$dx\_neg\_bin\_26$>$, dy:~$<$dy\_pos\_bin\_14$>$, dyaw:~$<$dyaw\_pos\_bin\_05$>$

\textbf{Predicted observation:}\quad\quad\quad\quad\quad\quad\quad\quad\quad~~~\includegraphics[height=1.7cm, width=2.265cm]{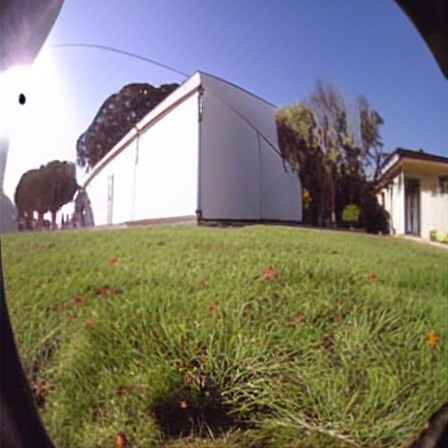}

\end{fullpromptbox}
\vspace{-0.2cm}
\caption{\small Prompt design example of LCVN-Uni (context size = 1).}
\label{fig:app_nv}

\end{figure*}

We examine detailed prompt formulation and response behaviors of coupling navigation planning and visualization in Fig.~\ref{fig:app_nv}. This example illustrates how multimodal inputs including navigation instructions and egocentric observations jointly guide both planner and world model in LCVN task.

\subsection{Training Loss Details for LCVN-Uni}

LCVN-Uni optimizes a unified objective that jointly learns navigation planning and world modeling through a combined loss function. At each training step, the model outputs logits for both action bins and visual tokens, enabling end-to-end optimization:
\begin{equation}
\mathcal{L}_{\text{joint}} = \mathcal{L}_{\text{plan}} + \lambda \mathcal{L}_{\text{imagine}},
\end{equation}
where $\mathcal{L}_{\text{plan}}$ is the discretized bin-token loss~\citep{dong2025unified} for action prediction, and $\mathcal{L}_{\text{imagine}}$~\citep{dong2025unified, li2025imagine} enforces fidelity in predicted future observations. The hyperparameter $\lambda$ balances the relative importance between planning and imagination.

Following the bin tokenization scheme, each navigation action $a_t \in \mathbb{R}^3$ is represented as $(x_t, y_t, \phi_t)$ and discretized into tokens drawn from disjoint sets $\mathcal{B}_x$, $\mathcal{B}_y$, and $\mathcal{B}_\phi$. Let $P(s_j)$ denote the predicted probability distribution over vocabulary tokens at position $j$. The action loss is defined as:
\begin{equation}
\!\!\!\!\mathcal{L}_{\text{plan}}\!\!=\!\!\frac{1}{3} \!\sum\nolimits_{d \in \{x,y,\phi\}}\!\! 
\Big(\!\!\!-\!\!\log P(b_d^*\!\mid\!s_j\!\in\!\mathcal{B}_d) \Big)\!\! +\! \mathcal{L}_{\text{text}},\!\!
\end{equation}
where $b_d^*$ is the ground-truth bin token for dimension $d$, and $\mathcal{L}_{\text{text}}$ is the cross-entropy loss for text tokens. For visual prediction, given ground-truth embedding $\mathbf{u}_j$ at token position $j$ (out of $m$ tokens in the predicted observation $\hat{o}_{t+1}$) and a codebook $\mathcal{C}=\{\mathbf{c}_1, \dots, \mathbf{c}_K\}$ of size $K$, imagination loss is:
\begin{equation}
\!\!\mathcal{L}_{\text{imagine}} \!=\! \frac{1}{m}\!\sum\nolimits_{j=1}^{m}\! 
\sum\nolimits_{k=1}^{K} \!\|\mathbf{u}_j - \mathbf{c}_k\|^2\! \cdot\! P(s_j \!= \!\mathbf{c}_k),\!\!
\end{equation}
where $\|\mathbf{u}_j - \mathbf{c}_k\|^2$ is the squared distance between $\mathbf{u}_j$ and codebook entry $\mathbf{c}_k$, and $P(s_j = \mathbf{c}_k)$ is the predicted probability of assigning token $j$ to codebook entry $k$. This joint optimization allows LCVN-Uni to learn shared multimodal representations that simultaneously support accurate navigation planning and visual imagination.

\section{Additional Experiments \& Results}
\label{sec:appx-exp-details}

\subsection{Evaluation Metric Details}
\label{appx.eval_metrics}

We evaluate overall system performance using two complementary categories of metrics in this work:

\noindent \textbf{Navigation Quality.} For the LCVN task, we report \textbf{Success Rate (SR)}~\citep{li2024human, dong2025ha}, which counts a trajectory as successful if its final distance $d$ to the target is smaller than the agent’s average step size $\bar{s}$. Formally, for $N$ trajectories with terminal estimate $\hat{p}^{(i)}_{T}$ and goal position $p^{(i)}_{g}$, SR is defined as:
\begin{equation}
\text{SR}=\frac{1}{N}\sum_{i=1}^{N}\mathbf{1}\!\left[d\!\left(\hat{p}^{(i)}_{T},\,p^{(i)}_{g}\right)<\bar{s}\right].
\end{equation}
We further report \textbf{Absolute Trajectory Error (ATE)}, which measures global accuracy by computing the Euclidean distance between aligned predicted and reference poses, and \textbf{Relative Pose Error (RPE)}, which captures local consistency by quantifying deviations in relative motion between consecutive estimated and ground-truth poses~\citep{sturm2012evaluating}.

\noindent \textbf{Imagination Quality.} To evaluate predicted visual rollouts, we employ both structural and perceptual measures: \textbf{SSIM}~\citep{wang2004ssim}, \textbf{PSNR}~\citep{hore2010image}, \textbf{LPIPS}~\citep{zhang2018unreasonable}, and \textbf{DreamSim}~\citep{fu2023dreamsim}. The latter two are deep perceptual metrics designed to better approximate human judgments. To capture long-horizon stability, we extend these metrics to rollout evaluation, reporting \textbf{SSIM$@n$}, \textbf{PSNR$@n$}, \textbf{LPIPS$@n$}, and \textbf{DreamSim$@n$}. One-step metrics compare the ground-truth next frame $o_{t+1}$ with the one-step prediction $\hat{o}^{(1)}_{t+1}$ obtained from $(o_t,a_{t+1})$. For horizon $n$, we perform open-loop rollouts by recursively feeding predicted frames back into the model while conditioning on the ground-truth action sequence $a_{t+1:t+n+1}$. For example, $\mathrm{SSIM}@n$ is calculated as:
\begin{equation}
\mathrm{SSIM}@n=\mathrm{SSIM}(o_{t+n},\,\hat{o}^{(n)}_{t+n}),
\end{equation}
with analogous definitions for PSNR$@n$, LPIPS$@n$, and DreamSim$@n$. We set $n$ as 8 for experiments.

\subsection{Implementation Details}
\label{sec:appx-impl} 

\noindent \textbf{Diamond}~\citep{alonso2024diffusion} We select Diamond as one of the world‑model baselines for comparison, as it represents a typical UNet‑based diffusion world model. We adopt DIAMOND in its best configuration following the official implementation~\citep{alonso2024diffusion}. The diffusion model is trained to autoregressively predict future states at a resolution of 56×56, with an additional upsampler to generate outputs at 224×224. Continuous actions are incorporated through a linear embedding layer for conditioning. We pair Diamond with LCVN‑AC to jointly support action prediction and imagination, training LCVN‑AC from scratch on LCVN dataset train split, with Diamond replacing LCVN‑WM as world model.

\noindent \textbf{NWM}~\citep{bar2025navigation}
We select NWM as another baseline world model for comparison, as it represents a typical DiT‑based diffusion architecture. For training, we adopt NWM’s best reported configuration: context length = 4, goal state = 4, and a CDiT‑XL backbone with 1B parameters, training NWM from scratch on the LCVN dataset train split. For inference, we follow the settings in their paper and use 250 diffusion denoising steps to achieve optimal performance. To jointly support action prediction and imagination, we pair NWM with LCVN‑AC, training LCVN‑AC from scratch on LCVN dataset train split, with NWM replacing LCVN‑WM as world model.

\noindent \textbf{LCVN-WM}
For LCVN‑WM, we first train an ImageVAE~\citep{kingma2013auto} from scratch to compress input images into 32×32 latent representations. Then we build the model upon a DiT‑XL backbone with approximately 1B parameters and a hidden size of 1152. Training leverages diffusion forcing~\citep{chen2024diffusion}, ensuring that each frame retains independent noise levels. We employ the AdamW optimizer~\citep{loshchilov2017decoupled} with linear warmup and a constant learning rate of 1e-4. For sampling and inference, we adopt the deterministic DDIM sampler~\citep{song2020denoising} with 50 steps following~\citep{song2025history}.

\noindent \textbf{LCVN-Uni}
LCVN-Uni is fine-tuned on GAIR Anole-7B~\citep{chern2024anole} with a 4096-token context, while keeping the bin, text and image tokenizers frozen. Input images are resized to $448 \times 448$ (height $\times$ width) and discretized into 784 visual tokens when the context size is 1 or 2, and into 625 tokens when the context size is 4. During training, only the LoRA~\citep{hu2022lora} adapters (rank = 16) in the Transformer’s \textit{qkv} projections are updated~\citep{liu2023llava}. Optimization is performed with AdamW for 20 epochs using a learning rate of $2 \times 10^{-4}$. Training is conducted on 4$\times$NVIDIA A100 GPUs (80GB each) with a global batch size of 8 (per-GPU batch size = 1, gradient accumulation = 2).

\subsection{More Experiments}

\noindent \textbf{Inference Speed.}~Table~\ref{tab:LCVN_variants} reports the average inference time per step, including both action prediction and imagination, evaluated on the LCVN val seen split. 
Inference efficiency is a critical factor for real-world deployment, as embodied agents must operate under strict latency constraints~\citep{deng2026hierarchical}. 
Compared with NWM~\citep{bar2025navigation} and LCVN-Uni, LCVN-WM achieves substantially lower inference time while maintaining competitive performance. 
Further acceleration is possible through reducing the diffusion denoising steps from 50 to 6 through model distillation~\citep{wang2024phased}. 
Quantization to 4-bit, which we have not yet explored, is expected to provide additional speedup without significant performance degradation~\citep{frantar2022gptq}. 
These results highlight LCVN-WM as a practical and efficient alternative for navigation tasks requiring both accuracy and real-time responsiveness.

\begin{table}[htbp]
\small
\centering
\caption{\small \textbf{Inference time (seconds)} per step, averaged on the LCVN dataset val seen split, including both action prediction and imagination. 
For NWM and LCVN-WM, the times include the latency of LCVN-AC’s action prediction.}
\resizebox{0.85\linewidth}{!}{
\begin{tabular}{ccccc}
\toprule
\textbf{NWM} & 
\textbf{LCVN-Uni} & 
\textbf{LCVN-WM} & 
\textbf{LCVN-WM (+Distillation.)} & 
\textbf{LCVN-WM (+Quant. 4-bit)} \\
\hline
 \rowcolor{w_1!7} 11.2 & 20.5 & 6.4 & 0.6 & 0.1 (est.~\citep{frantar2022gptq})\\
\bottomrule
\end{tabular}}

\label{tab:LCVN_variants}
\end{table}

\noindent \textbf{Interleave \emph{vs.} Predict Both.}~Table~\ref{tab:predict_both_comparison} compares two step strategies for LCVN-Uni: Interleave, where action and observation are predicted in alternating substeps, and Predict Both, where both are jointly predicted in a single forward pass. Predict Both achieves comparable navigation and imagination accuracy while offering markedly higher computational efficiency (1.3$\times$), which empirically supports our design choice to adopt joint prediction of actions and observations in the LCVN-Uni agent. 

\begin{table}[!t] 
  \tabcolsep=0.18cm
 \setlength{\abovecaptionskip}{3pt} 
 \setlength{\belowcaptionskip}{1pt} 
     \centering 
        \caption{\small \textbf{Interleave~vs.~Predict Both} on navigation and imagination performance of LCVN-Uni ($k$=2) on LCVN val seen.} 
 
     \resizebox{\linewidth}{!}{%
     \begin{tabular}{p{3.2cm}p{1cm}p{1cm}<{\centering}p{1cm}<{\centering}p{1cm}<{\centering}p{1cm}<{\centering}p{1.8cm}<{\centering}p{1.8cm}<{\centering}p{2.5cm}<{\centering}} 
     \hline 
     & \multicolumn{4}{c}{\textbf{Navigation}} & \multicolumn{4}{c}{\textbf{Imagination}} \\ 
     \cmidrule(lr){2-5} \cmidrule(lr){6-9} 
     \textbf{Method} 
     & \textbf{Speed} & \textbf{ATE$\downarrow$} & \textbf{RPE$\downarrow$} & \textbf{SR$\uparrow$} 
     & \textbf{SSIM$\uparrow$} & \textbf{DreamSIM$\downarrow$} 
     & \textbf{SSIM$@8$$\uparrow$} & \textbf{DreamSIM$@8$$\downarrow$} \\ 
     \hline 
     LCVN-Uni (Inter.)& 1$\times$ 
     & \textbf{0.34} & 0.12 & \textbf{0.44} 
     & \textbf{0.438} & 0.074 
     & 0.201 & \textbf{0.115} \\ 
     \rowcolor{w_1!21} \textbf{LCVN-Uni}  & \textbf{1.3$\times$} 
     & 0.36 & \textbf{0.11} & 0.42 
     & 0.423 & \textbf{0.072} 
     & \textbf{0.218} & 0.119\\ 
     \hline 
     \end{tabular}} 

     \label{tab:predict_both_comparison} 
        
\end{table}

\noindent \textbf{Generalization in Unseen Environments.}~Table~1~highlights~that LCVN-Uni demonstrates stronger navigation performance in unseen environments, consistently outperforming LCVN-WM + LCVN-AC in SR while maintaining competitive ATE and RPE. This advantage arises from its unified architecture, which jointly models navigation planning and world modeling, enabling shared representations that transfer more effectively to novel scenarios.~In contrast, Table~2 shows that LCVN-WM achieves superior imagination performance, indicating that while the imagination ability indeed influences navigation outcomes (Sec.~5.2), coupling of perceptual alignment with planning is even more critical for sustaining semantic consistency under distributional shifts.~These findings underscore the importance of both imagination capacity and representation sharing as key drivers of generalization for world models in LCVN.

\section{Limitations}
\label{sec:limitations}

While LCVN provides a controlled testbed for studying language, imagination, and decision-making in embodied agents, several limitations remain. First, following prior goal-conditioned navigation work~\citep{shahvint, sridhar2024nomad, shah2022gnm}, we adopt an offline trajectory-prediction protocol, which enables reproducible comparison across heterogeneous real-robot data but does not capture closed-loop behaviors such as recovery from execution errors or interaction with dynamic agents; closed-loop validation on physical platforms is a natural complement. Second, all instructions in the LCVN dataset are written in English and follow three pre-defined styles (Concise, Intricate, Landmark-based); extending the benchmark to multilingual instructions and to noisier, more naturalistic phrasings that arise in real human–robot interaction is a promising direction. Third, our zero-shot evaluation relies primarily on TartanDrive's off-road driving scenes as the unseen split, which provides a strong distribution shift in terrain and embodiment but leaves additional axes of generalization open for future exploration.

\section{Broader Impacts}
\label{sec:broader_impacts}

LCVN targets language-conditioned visual navigation, which has natural applications in assistive robotics, service robots, and accessible human–robot interfaces, where users may prefer natural language over visual goals or low-level commands. We expect direct negative societal impacts to be limited: the world models studied in this work generate short-horizon egocentric navigation imagery rather than photorealistic media of identifiable individuals, and the underlying datasets are publicly released robotic trajectories without personally identifiable content. As with other navigation and generative-modeling research, downstream misuse in surveillance or in fabricating first-person footage is conceivable; we encourage subsequent work building on LCVN to follow responsible release practices and to consider provenance and access controls when deploying derived systems.

\clearpage\clearpage
\bibliographystyle{assets/plainnat}
\bibliography{main}

\end{document}